\documentclass[]{uiuc_paper}
\pdfoutput=1

\usepackage[toc,page,header]{appendix}
\usepackage{cleveref}
\usepackage{subcaption}
\usepackage{booktabs}
\usepackage{tabularx}
\usepackage{colortbl}
\usepackage{algorithm}
\usepackage[noend]{algpseudocode}

\usepackage{latexsym}
\usepackage{url}
\usepackage{amssymb}
\usepackage{amsmath}
\usepackage{amsfonts}
\usepackage[utf8]{inputenc}
\usepackage{booktabs}
\usepackage{pifont}
\usepackage{multirow}
\usepackage{makecell}
\usepackage{paralist}
\usepackage{xspace}
\usepackage{adjustbox}

\crefname{section}{Sec.}{Secs.}\Crefname{section}{Sec.}{Secs.}
\crefname{subsection}{Sec.}{Secs.}\Crefname{subsection}{Sec.}{Secs.}
\crefname{subsubsection}{Sec.}{Secs.}\Crefname{subsubsection}{Sec.}{Secs.}
\crefname{appendix}{App.}{Apps.}\Crefname{appendix}{App.}{Apps.}
\crefname{subappendix}{App.}{Apps.}\Crefname{subappendix}{App.}{Apps.}
\crefname{subsubappendix}{App.}{Apps.}\Crefname{subsubappendix}{App.}{Apps.}
\crefname{figure}{Fig.}{Figs.}\Crefname{figure}{Fig.}{Figs.}
\crefname{table}{Tab.}{Tabs.}\Crefname{table}{Tab.}{Tabs.}
\crefname{equation}{Eq.}{Eqs.}\Crefname{equation}{Eq.}{Eqs.}
\crefname{algorithm}{Alg.}{Algs.}\Crefname{algorithm}{Alg.}{Algs.}

\newcommand{\method}{Keel}

\definecolor{wsNavy}{HTML}{2E7D57}%
\definecolor{wsBlue}{HTML}{6E86AE}
\definecolor{wsGrey}{HTML}{7C8591}
\definecolor{wsFog}{HTML}{BDD7EE}
\definecolor{wsIce}{HTML}{EEF2F7}
\definecolor{wsMist}{HTML}{E3EAF3}
\definecolor{wsRose}{HTML}{C4847F}
\definecolor{wsInk}{HTML}{22303C}
\definecolor{ours}{HTML}{2E7D57}

\title{Keel: A Frontier Model Teaches a Vision-Language Model by Rewriting Its Weights}

\author[1]{Jiajun Fan}
\author[1,4]{Neil He}
\author[1]{Bangji Yang}
\author[1]{Qinglong Zheng}
\author[1,2]{Ziwen Wang}
\author[1]{Chaoran Cheng}
\author[3]{Changyi Yang}
\author[1]{Hanqi Mao}
\author[1]{Ge Liu}

\affiliation[1]{\affillogo{\includegraphics{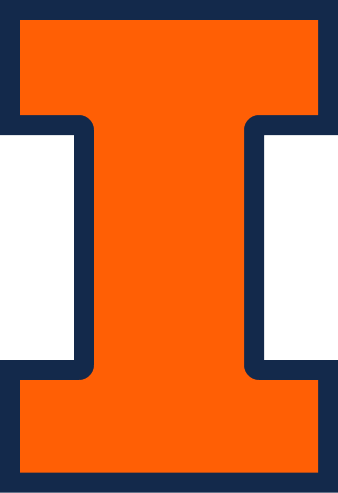}}{UIUC}}
\affiliation[2]{\affillogo{\includegraphics{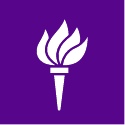}}{NYU}}
\affiliation[3]{\affillogo{\includegraphics{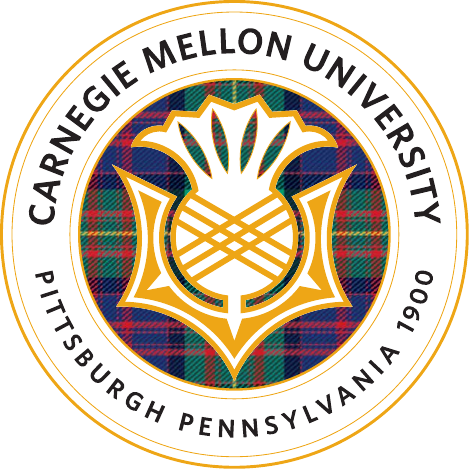}}{CMU}}
\affiliation[4]{\affillogo{\includegraphics{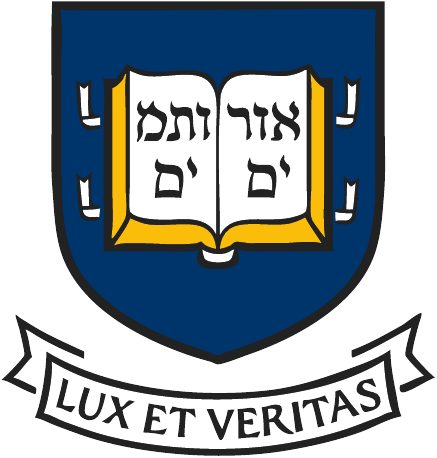}}{Yale}}

\abstract{Can a frontier model repair a vision-language model (VLM) by rewriting its weights? Small reasoning VLMs such as
Qwen3.5 need such a repair: in thinking mode they fall into loops on many image questions, repeating themselves until
the token limit, spending the whole budget and losing the answer. The usual remedies act on the decoding, leaving the
looping model as it was, or must be told what to change. Yet the cause is written in the weights. \method{}, the first
agentic weight-editing method for VLMs on image tasks, finds and removes it there. Told only the symptom,
an editor, a frontier multimodal LLM acting as an agent, locates the cause, edits the weights and tests them, with a failure census and instruments showing it the failing
images and where the VLM attends; among several editors, the harness keeps only edits that clearly reduce the failure
without lowering accuracy on questions no editor saw. The repair needs no training, keeps the model's size and
per-token cost, and applies to already-edited weights. The editors expose where the loops live: from the original
weights, independent editors on both models trace them to the attention outputs of the middle layers, and the smaller
VLM stops consulting the image inside a loop. On Qwen3.5-0.8B and Qwen3.5-2B, across six image benchmarks from visual
mathematics to GUI grounding, the edit removes most loops on every benchmark and shortens answers, and the edited
models are significantly more accurate than the thinking models on average; the larger one gains most on GUI
grounding, a task absent from the practice questions. Because the repair lives in the weights, it combines with budget
forcing where held-out questions favor it: on the best available weights, \method{} has the best average accuracy on
both models, significantly above every training-free baseline we compare on the larger one.
}

\date{\today}

\begin{document}

\maketitle

\begin{figure}[t]
\centering
\includegraphics[width=\linewidth]{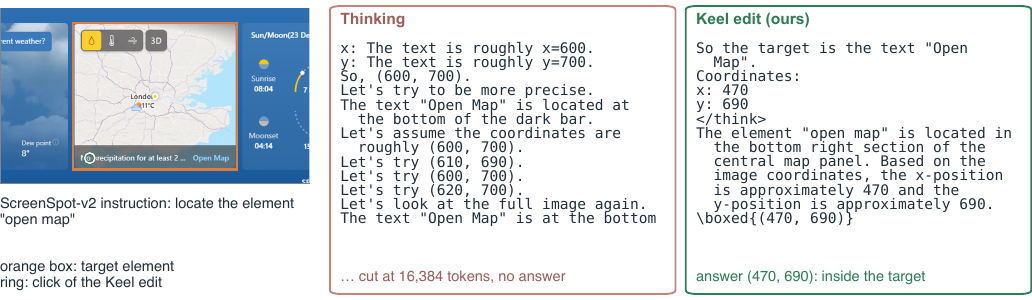}
\caption{\textbf{The edit stops a thinking loop.} Qwen3.5-0.8B repeats the same estimates until it is cut at
the token limit (middle); after the \method{} edit, the same model closes its thinking and clicks inside the target
(right). Selection rule in \cref{app:qual-rule}.}
\label{fig:teaser}
\end{figure}
\begin{figure}[t]
\centering
\includegraphics[width=\linewidth]{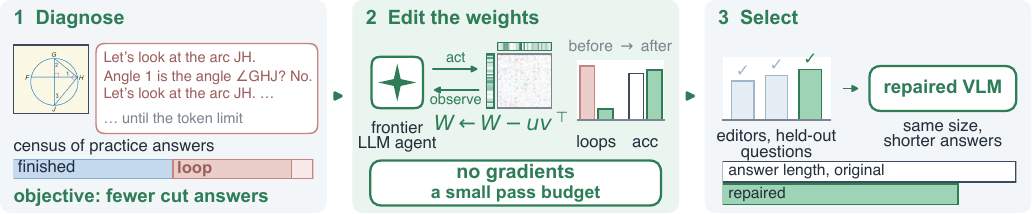}
\caption{\textbf{\method{} repairs a VLM by rewriting its weights, without gradients} (real data of the 0.8B run from the original weights).
(1) The census shows how the VLM fails; the repair objective targets it. (2) The editor, a frontier LLM agent, edits the weights turn by turn: it acts (code, forward passes) and observes the results.
(3) The edits of several editors are tested on held-out questions; the best is kept.}
\label{fig:pipeline}
\end{figure}

\section{Introduction}
\label{sec:intro}

\begin{figure}[t]
\centering
\includegraphics[width=\linewidth]{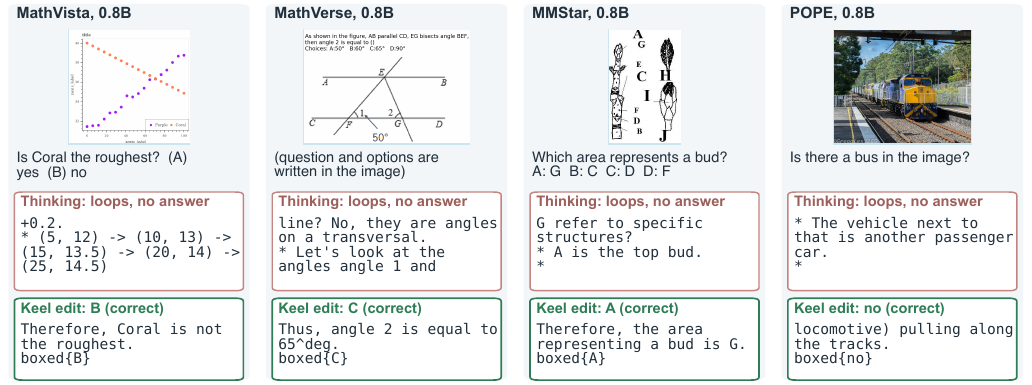}
\caption{\textbf{The edit stops the loops on four more benchmarks.} Qwen3.5-0.8B, one question per benchmark
(rule in \cref{app:qual-rule}); rose: the thinking model repeats itself until it is cut without an answer; green: the end of the
edited model's answer, which is correct.}
\label{fig:qual}
\end{figure}

Small vision-language models (VLMs) now think before they answer, and on image tasks the thinking often does not end.
The model card of Qwen3.5~\cite{qwen3.5} warns that its smallest models are more prone to entering thinking loops, and
on image benchmarks the warning describes a major failure: with the card's own sampling settings, a large share of
the thinking model's answers never leave the thinking phase, and most of those end in text that repeats itself until
the tokens run out (\cref{fig:teaser,fig:qual}). Each such answer spends the whole token budget and is lost, however good the
reasoning before the loop was, while the answers that do finish are mostly correct (\cref{sec:method}). Loops are
most frequent on the vision-only problems of MathVerse and on GUI grounding, the step on which visual agents act
(\cref{sec:main}).

The remedies in use act on the decoding and leave the model that loops as it was. Loops, however, arise from errors
in what the model has learned, and sampling differently reduces them without removing that
cause~\cite{pipis2025wait}. Penalising repeated tokens and banning reflection words change which words the model may
write, not what drives it into a loop. Budget forcing asks for an answer at a budget but does not keep the model from
running to it: about as many answers reach the budget as the thinking model loses at the token limit (\cref{app:loops}).
Directional ablation~\cite{arditi2024refusal} edits the weights, but by a fixed recipe, and on the larger model no
edit it finds passes its own accuracy check. What is missing is a way to find the cause in the network and remove it there.

\method{} repairs the model itself, the way an engineer repairs code: the editor, a frontier multimodal LLM acting as an agent, reads the VLM's
weights, rewrites the faulty part and tests the fix, without gradients or training. The result is plain weights of
unchanged size and per-token cost, and the repair also applies to already-edited weights. Interpretable weight-editing
methods~\cite{meng2022locating,arditi2024refusal,sun2025thinkedit} must be told in advance what to change, a fact to
rewrite or a contrast that defines the direction to remove; \method{} names only the symptom and leaves finding the
mechanism, choosing where to act and writing the fix to the editor. A recent blog post had the same frontier model, shown no
images, remove a loop direction from the language model of Qwen3.5-0.8B on text-only
mathematics~\cite{takehi2026weightspace}; \method{} is the first to bring such agentic editing to VLMs on image
tasks, with editors that see the images and edits selected on questions no editor saw.

An editor is only as good as the signal it is judged on. Within the forward passes an editor can afford, a real
accuracy gain of a few points cannot be told from noise, while the share of looping answers moves far beyond it once
a repair works; editors given no diagnosis and asked only to raise accuracy mostly commit no edit
(\cref{app:editors}). \method{} therefore changes what the editor is asked to do and who decides (\cref{fig:pipeline}). A
\emph{failure census} of the model on practice images tells the editor how it fails. A \emph{repair objective} states the goal as
removing that failure without a significant drop in accuracy, and the harness holds every final edit to it. A \emph{selection step} runs several editors
independently; the harness, not the editors, keeps the most accurate edit that clearly reduces the failure without
losing accuracy on held-out questions. Visual instruments show the editor the failing images, where the VLM's
attention falls on them, and what blanking the image changes.

The editors also show where the loops live: from the original weights, independent editors on both models return to
the attention outputs of the middle layers; removing the 0.8B editor's directions there more strongly removes more
loops and amplifying them adds loops; and inside a loop the 0.8B model's writing no longer depends on the image. Removing the loops pays off in
accuracy. On Qwen3.5-0.8B and Qwen3.5-2B, across six image benchmarks from visual mathematics to GUI grounding, the
edit removes most loops on every benchmark, shortens the answers and makes the model significantly more accurate on
average; the larger model gains most on GUI grounding, a task absent from the practice questions. Because the edit
acts on the model and budget forcing on the decoding, the two combine: with the edit, fewer answers reach the budget
on every benchmark. On the best available weights (2B: the original model; 0.8B: the blog post's
edit~\cite{takehi2026weightspace}, which \method{} repairs further), \method{} has the best average accuracy on both models, significantly
above the thinking model, every decoding-time baseline and, on 0.8B, the directional ablation
(\cref{fig:overview}).

\noindent Our contributions are:
\begin{itemize}
\item \method{}, the first agentic weight-editing method for VLMs on image tasks: told only the symptom, a frontier
LLM agent that sees the images and where the VLM looks finds the cause in its weights and rewrites it without training.
\item Where loops live: from the original weights, independent editors on both models trace thinking loops to the attention
outputs of the middle layers, and the smaller model stops consulting the image inside a loop.
\item A design that makes editing a reliable repair, shown on 0.8B: a failure census, a repair objective and
held-out selection; each ablation leaves significantly more loops.
\item Gains as a consequence: both models loop far less and become significantly more accurate, and on the best available
weights \method{} has the best average accuracy on both.
\end{itemize}

\section{Method}
\label{sec:method}

\method{} has four parts (\cref{fig:pipeline}; pseudocode in \cref{app:procedure}): a failure census, a repair objective, an editing harness with visual
instruments, and selection among several independent editors.

\subsection{Problem Formulation}
\label{sec:problem}
A VLM $f_\theta$ with weights $\theta_0$ answers an image $v$ and a question $q$ with $f_\theta(v,q;\pi)$
under a fixed inference protocol $\pi$ (chat template, image preprocessing, decoding). A practice pool
$\mathcal{D}_{\mathrm{prac}}=\{(v_i,q_i,a_i)\}$ with reference answers $a_i$ contains no evaluation question. An \emph{editor} $E$, a frontier multimodal LLM acting as an agent, returns weights $\theta\in\Theta(\theta_0)$
with the tensor names and shapes of $\theta_0$. The goal is higher accuracy $\mathrm{Acc}(\theta;\mathcal{D}_{\mathrm{test}})$
on benchmarks $\mathcal{D}_{\mathrm{test}}$ that $E$ never sees, under two constraints. (i)~\emph{No training}: $E$ may
change $\theta$ in any way, but not by gradients through the network, optimisers or iterative fitting
of weights to a loss. (ii)~\emph{Limited observation}: $E$ observes $f_\theta$ only through forward passes on
practice questions, at most $B$ of them on its own request.

\subsection{Failure Census}
Before any editor is started, the harness runs the unedited model once on 512 practice questions under the
inference protocol and summarises how it fails: accuracy, the share of answers cut at the token limit, the share of
those cut answers that end in repetitive text, and the accuracy of the answers
that did finish. An answer counts as repetitive when its last 6{,}000 characters compress to less than 15\% of their
size; a cut answer with a repetitive ending is a \emph{verbatim loop}. On both models we study, the census points to the same failure: a large share of answers never close their
thinking, most of them ending in repetitive text, while the answers that do finish are mostly correct
(\cref{app:census}). The census and all 512 answers are given to the editor as its starting diagnosis, with the note that
repairing a diagnosed failure such as repetition is in scope.

\subsection{Repair Objective}
\label{sec:objective}
An editor judged only by accuracy faces a statistical problem. A useful edit may raise accuracy by a few points,
while a generation pass of a few hundred questions measures accuracy with a standard error (SE) of about the same size;
within its budget the editor cannot tell a real gain of this size from noise, so it can discard a candidate whose
gain is real. The diagnosed failure gives a far clearer signal: a working repair moves the share of answers that end in a loop by
many standard errors (\cref{app:objective}). Whether an answer is cut at the limit, and whether it
ends in a loop, is observed on every question, and it can be seen early: a short pass shows the onset of repetition
at a fraction of the cost of a full one. \method{} therefore states the task as a repair: the editor must
(a)~clearly reduce the share of answers lost at the token limit, the failure the census diagnosed, and (b)~keep
accuracy not significantly lower than the original. Under the repair objective, every paired generation pass also
reports the paired difference in the share of cut answers with its standard error, so the editor can test (a)
directly.

\subsection{Editing Harness}
\label{sec:harness}
The editor works as an agent in turns, replying to each result with an explanation and exactly one action,
either Python code or a forward-pass request. A session runs until its pass budget is used (or 260 turns or 14 hours have passed) and ends
with the editor's final choice; long sessions continue in a fresh conversation every 40 turns (\cref{app:procedure}).

\noindent\textbf{Kernel.} Code runs in a persistent CPU-only Python kernel that holds the checkpoint and exposes
functions to list tensors, to load a tensor from the current edit or from the original, to save or drop an edit,
the tokenizer, the practice pool, and loaders for the results of earlier forward passes. Autograd and optimisers
are disabled, and code that tries to train, to run the network, or to use it in any other way is refused. The input
embedding, the output layer and the final norm are out of scope, as is any edit that only rescales a whole tensor.
The kernel never holds a computational graph of the network: the editor can compute anything from the weights and
from the outputs of the forward passes it requested, but it cannot backpropagate through the network or run it. This
enforces constraints (i) and (ii) of \cref{sec:problem} instead of leaving them to instructions.

\noindent\textbf{Forward passes.} Each pass counts against the budget; a failed request does not. A
\emph{generation pass} has the VLM answer up to 512 practice questions with the inference protocol, using the
original weights or the current edit, or up to 256 with both on the same questions and per-question sampling seeds.
Every summary reports, for each model, accuracy, the share of cut answers (for editors with the census also the share
of cut answers that end in repetition), the share without an answer and the mean length; the paired case adds the accuracy difference with its standard
error and the counts of answers that flipped each way, and, under the repair objective, the paired difference in the
share of cut answers. A \emph{teacher-forced read} feeds up to 256 (question, answer text) pairs
and returns hidden states after any layer, MLP activations and token log-probabilities, reduced over positions
chosen by the editor.

\noindent\textbf{Visual instruments.} Because the editor is itself multimodal, the harness lets it look. The editor
can attach practice images, optionally with the VLM's answer, to its next message, as well as figures it drew
itself. A teacher-forced read can replace an image by a uniform gray image to measure what the image contributes (a
\emph{blank-image read}); it can report, for every full-attention head and token, the share of attention that lands on image tokens; and it can
return attention maps over the image grid at chosen answer positions.

\subsection{Selection Among Editors}
\label{sec:select}
A single editing session is not reliable: the same task can end with an edit that passes the repair test below or
with no edit at all. We
therefore run several editors independently and let the harness, not the editors, choose. Each editor's final edit
is scored by a paired comparison with $\theta_0$ on 1{,}000 held-out practice questions that no editor saw, with the
same sampling seeds for both models. An edit passes this \emph{repair test} if it reduces the share of cut answers
by more than twice its SE and its accuracy difference is positive. Among passing edits the most accurate is kept. If the two
best passing edits touch disjoint tensors, their union is scored as well and kept if it passes and is more accurate.

\subsection{Deployment}
\label{sec:deploy}
The edit acts on the model and decoding-time fixes act on the generation, so the two can be combined. \method{}
applies the edit made on the starting weights that are most accurate on held-out questions, and adds budget
forcing~\cite{muennighoff2025s}, which closes any thinking that reaches the budget and asks for the answer, if that is
more accurate on them.

\section{Experiments}
\label{sec:exp}
\begin{figure}[t]
\centering
\includegraphics[width=0.62\linewidth]{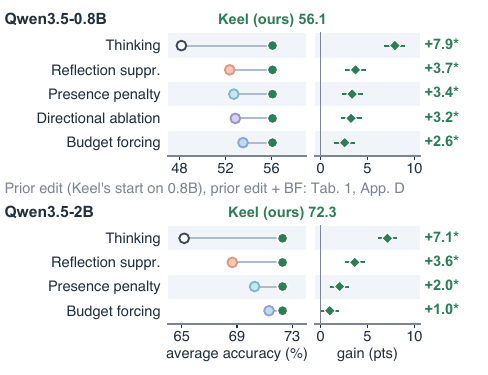}
\caption{\textbf{\method{} is the most accurate method on both models.} Average accuracy over the six benchmarks of
the baselines and of \method{} (green), and \method{}'s paired gain over each with its 95\% interval ($^\ast$:
excludes zero). Higher is better.}
\label{fig:overview}
\end{figure}

\begin{table}[!t]
\centering
\footnotesize
\setlength{\tabcolsep}{5.2pt}
\adjustbox{max width=\linewidth}{\begin{tabular}{lcccccccc}
\toprule
Method & MathVista\,$\uparrow$ & MathVerse\,$\uparrow$ & MMStar\,$\uparrow$ & HallBench\,$\uparrow$ & POPE\,$\uparrow$ & ScreenSpot\,$\uparrow$ & Avg.\,$\uparrow$ & Gain\,$\uparrow$ \\
\midrule
\multicolumn{9}{l}{\textit{Qwen3.5-0.8B}} \\
Thinking & 50.6 & 21.3 & 47.3 & 61.1 & 86.5 & 22.2 & 48.2 & -- \\
Presence penalty & 53.8 & 26.1 & 51.7 & 66.0 & 86.9 & 31.8 & 52.7 & +4.6 \\
Reflection suppression & 55.4 & 27.8 & 50.5 & 65.6 & 86.8 & 28.1 & 52.4 & +4.2 \\
Budget forcing & 54.4 & 23.1 & 49.9 & 69.0 & 87.9 & 36.8 & 53.5 & +5.4 \\
Directional ablation & 55.3 & 25.6 & 51.3 & 66.6 & 87.1 & 31.2 & 52.9 & +4.7 \\
Prior edit~\cite{takehi2026weightspace} & 58.7 & 28.4 & 53.4 & 64.7 & 86.9 & 38.0 & 55.0 & +6.9 \\
Prior edit + budget forcing & 57.6 & 30.7 & 53.3 & 67.0 & 87.1 & 37.5 & 55.5 & +7.4 \\
\rowcolor{ours!12}\method{} (ours) & 59.5 & 30.8 & 57.4 & 65.7 & 86.7 & 36.4 & \textbf{56.1} & \textcolor{wsNavy}{\textbf{+7.9}} \\
\midrule
\multicolumn{9}{l}{\textit{Qwen3.5-2B}} \\
Thinking & 70.9 & 45.3 & 61.5 & 68.1 & 85.8 & 59.4 & 65.2 & -- \\
Presence penalty & 72.7 & 51.5 & 64.7 & 71.8 & 88.1 & 72.9 & 70.3 & +5.1 \\
Reflection suppression & 72.1 & 47.8 & 63.6 & 70.9 & 86.8 & 70.8 & 68.7 & +3.5 \\
Budget forcing & 74.2 & 51.1 & 64.0 & 73.4 & 87.5 & 77.7 & 71.3 & +6.2 \\
\rowcolor{ours!12}\method{} (ours) & 74.5 & 53.7 & 66.7 & 74.0 & 86.8 & 78.1 & \textbf{72.3}$^\dagger$ & \textcolor{wsNavy}{\textbf{+7.1}} \\
\bottomrule
\end{tabular}
}
\caption{\textbf{\method{} has the best average accuracy on both models.} Accuracy (\%); best average in
bold; $^\dagger$: significantly above every baseline (paired bootstrap, 95\%). \method{}: the repair edit
on the held-out-best starting weights (0.8B: the prior edit~\cite{takehi2026weightspace}; 2B: the original), with budget
forcing on 2B only, chosen on held-out questions. Gain: over Thinking, from unrounded averages. On 2B directional ablation kept no edit.}
\label{tab:main}

\vspace{8pt}
\begin{minipage}[t]{0.485\linewidth}
\centering
\includegraphics[width=\linewidth]{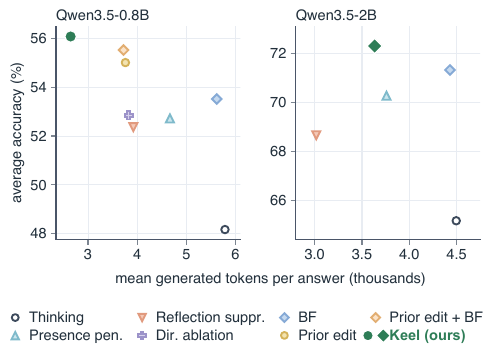}
\captionof{figure}{\textbf{\method{} is the most accurate method and shorter than the strongest baseline.} Average accuracy
and generated tokens per answer, every method of \cref{tab:main}; diamonds: with
budget forcing (BF); on 2B, directional ablation kept no edit (not shown). Higher accuracy and fewer tokens are
better.}
\label{fig:cost}
\end{minipage}\hfill
\begin{minipage}[t]{0.485\linewidth}
\centering
\includegraphics[width=\linewidth]{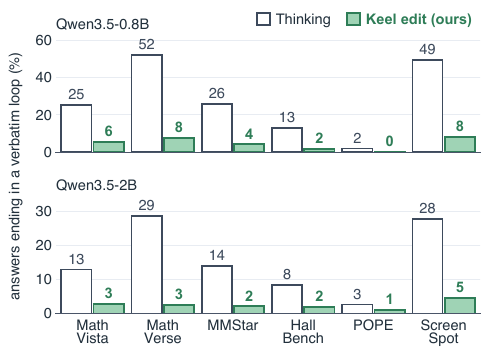}
\captionof{figure}{\textbf{The edit removes most verbatim loops on every benchmark.} Share of all answers that are cut at the
token limit while ending in repetitive text, for the thinking model and for the \method{} edit. Lower is better.}
\label{fig:loops}
\end{minipage}
\end{table}

\subsection{Experimental Setup}
\label{sec:setup}
\begin{figure}[t]
\centering
\includegraphics[width=\linewidth]{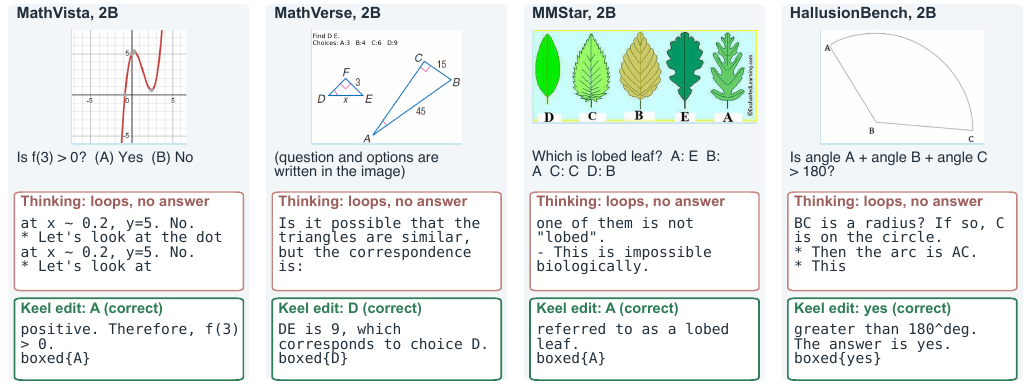}
\caption{\textbf{The edit also stops the loops on Qwen3.5-2B.} One question per benchmark (rule in \cref{app:qual-rule}); rose:
the thinking model repeats itself until it is cut without an answer; green: the end of the edited model's answer,
which is correct.}
\label{fig:qual2b}
\end{figure}

\noindent\textbf{Models.} We edit Qwen3.5-0.8B and Qwen3.5-2B~\cite{qwen3.5}, two natively multimodal models with
the same layout: a ViT encoder on $16{\times}16$ patches whose $2{\times}2$-merged patches are projected into a
24-layer language model, in which layers 3, 7, 11, 15, 19 and 23 use full attention with a gated output and the
other layers are Gated DeltaNet linear-attention layers. The language model has width 1{,}024 in the 0.8B model and
2{,}048 in the 2B model.

\noindent\textbf{Inference protocol.} The same protocol is used for practice passes, whose token limit an editor may
lower, and for evaluation: thinking
mode on, the model card's sampling for vision tasks (temperature 0.6, top-$p$ 0.95, top-$k$ 20), at most 16{,}384
new tokens, and images resized to at most 1{,}310{,}720 pixels. The final answer is the last
\texttt{\textbackslash boxed\{\}} after the thinking; an answer that never closes its thinking is wrong.

\noindent\textbf{Editors.} Each editor is GPT-6 Astra with maximal reasoning effort, accessed through an API, with a
budget of 40 forward passes (25 without the census). The practice pool is ViRL39K~\cite{wang2025vlrethinker}, a collection of verifiable
visual questions on geometry, charts, science and spatial reasoning; we remove multi-image items, items that repeat
an evaluation question and items that a difference-hash screen flags as near-duplicates of an evaluation image
(\cref{app:pool}), and hold out 1{,}000 items for selection. From the original weights, \method{} selects among three
visual editors with the census and the repair objective on the 0.8B model, and on the 2B model among four visual
editors with the census, two of which also receive the objective; there the selected edit is the union of the edits
of the two editors without it (\cref{app:editors} lists every editor and why the pools differ). We call this edit the \emph{\method{} edit}.
\emph{\method{}} itself is deployed as in \cref{sec:deploy}: on the 0.8B model on the prior edit (below), on which
three further editors run (\cref{app:blog}), without budget forcing; on 2B on the original weights, with it.

\noindent\textbf{Benchmarks.} Six image benchmarks, all of which need the image: visual mathematics in
MathVista-mini~\cite{lu2023mathvista} and the vision-only split of MathVerse-mini~\cite{zhang2024mathverse}, whose
questions are written inside the image; multimodal reasoning in MMStar~\cite{chen2024are}; hallucination in
HallusionBench~\cite{guan2023hallusionbench} (HallBench); object probing in POPE~\cite{li2023evaluating} (a fixed
random subset of 500 questions per split); and GUI element grounding for visual agents in
ScreenSpot-v2~\cite{cheng2024seeclick,wu2024osatlas}. We draw one sample per question. Differences are tested
with a paired bootstrap over questions (10{,}000 resamples); for the average over the six benchmarks, each
benchmark is resampled separately.

\noindent\textbf{Baselines.} All baselines are training-free and act on the same model. \emph{Thinking} is the
model as released. \emph{Presence penalty}
applies the card's suggested remedy for endless repetition (1.5). \emph{Reflection suppression} bans the tokens of
reflection words during decoding, following~\cite{wang2025wait}. \emph{Budget forcing}~\cite{muennighoff2025s}
closes the thinking once it reaches $16{,}384-1{,}024$ tokens with the model's own thinking-budget sentence, after
which the model answers in at most 1{,}024 tokens. \emph{Directional ablation}~\cite{arditi2024refusal} is the
automated counterpart of an edit: it projects the difference between the mean residual states of failing and
correct practice answers out of every matrix that writes to the residual stream, with layer and strength chosen by
paired passes on practice questions, within the forward-pass budget of an editor. \emph{Prior edit} is the 0.8B edit
released with the blog post~\cite{takehi2026weightspace}.

\subsection{Main Results}
\label{sec:main}

\noindent\textbf{\method{} has the best average accuracy on both models.} \method{} (ours) is significantly more accurate
on average than the thinking model and every decoding-time baseline on both models, and the directional ablation on 0.8B
(\cref{fig:overview}; per benchmark: \cref{tab:main}), and writes significantly shorter answers than the
strongest baseline of each model, on every benchmark (\cref{fig:cost}; \cref{app:cost}).

\noindent\textbf{The edit alone repairs the reasoning model.} On both models, the \method{} edit is significantly
more accurate than the thinking model it starts from, on average and on most single benchmarks: MMStar, HallusionBench and
ScreenSpot-v2 on both models, MathVista on the smaller one and POPE on the larger one. On GUI grounding the edit wins
back the most answers from the token limit on both models, and the larger model gains most there (\cref{app:loops}); the
gain comes with significantly fewer loops on every benchmark of both models (\cref{fig:loops}). Applying the edit step a second time, to the repaired 0.8B model, improves it
significantly further with an edit of six numbers (\cref{app:round2}).

\noindent\textbf{The edit combines with budget forcing.} Budget forcing ends most loops rather than preventing
them (\cref{app:loops}); the edit acts on the model, so the two can be used together: with the edit, significantly fewer answers reach the budget and
have to be forced to an end, on every benchmark and both models (\cref{app:loops}).

\noindent\textbf{The repair is cheaper at inference, and transfers.} The edited model is not only more accurate but
also shorter: on both models it generates significantly fewer tokens per answer than the thinking model, because
far fewer answers spin until the token limit (\cref{app:cost}). The practice questions are
geometry, chart, science and spatial-reasoning problems, none of them GUI grounding or object probing; the edits
nevertheless significantly improve GUI grounding on both models and object probing on 2B.

\subsection{Qualitative Results}
\label{sec:qual}

On single questions the failure looks the same on every benchmark, and the edit removes it
(\cref{fig:teaser,fig:qual,fig:qual2b}; one question per benchmark by a fixed rule, \cref{app:qual-rule}): the thinking model repeats one
step until it is cut at the token limit with no answer. After the edit, the same model
closes its thinking and answers correctly.

\begin{figure}[t]
\centering
\includegraphics[width=0.62\linewidth]{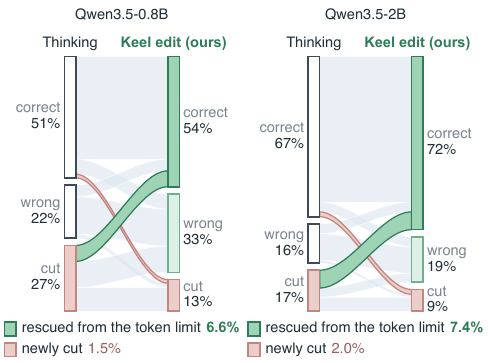}
\caption{\textbf{Question by question, the edit rescues answers lost at the token limit.} Outcome of every question
for the thinking model (left) and the \method{} edit (right), pooled over the six benchmarks.}
\label{fig:flow}
\end{figure}

\noindent\textbf{Where the gain comes from.} Following every question from the thinking model to the \method{} edit (\cref{fig:flow}) shows
where the accuracy comes from: the edit rescues several times more answers from the token limit than it newly cuts. Most of the
answers still cut after the edit are long thinking that does not repeat itself, which is what budget forcing
handles.

\subsection{What the Editors Found}
\label{sec:analysis}

\noindent\textbf{The repair objective makes progress visible.} The paired generation passes of the editor whose edit
\method{} selects from the original 0.8B weights are traced in \cref{app:mech}. It first screened single MLP neurons, loop directions in a few layers and one
attention head with cheap short passes and saw no clear or lasting effect on loops. It then removed a loop direction at
every middle layer, and from then on most passes show the loop share dropping far beyond
noise while it tunes where and how strongly to remove them; the passes in which loops rise are its tests of
removing them from the MLPs alone and of amplifying them. The repair objective
lets the editor commit the edit on its loop reduction rather than on an accuracy change no single
pass can resolve.

\begin{figure}[t]
\centering
\includegraphics[width=0.62\linewidth]{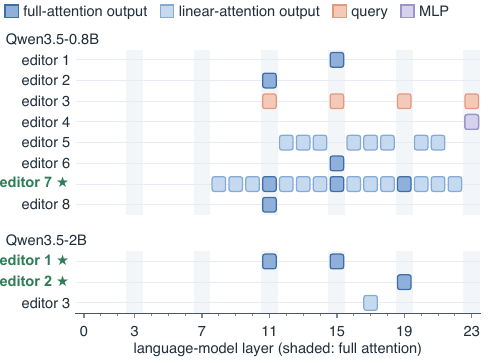}
\caption{\textbf{Most editors change attention output projections in the middle layers.} Every committed edit of
the editing runs that start from the original weights, on both models, one row per editor; a cell marks a changed matrix in that layer. $\star$: the \method{} edits.}
\label{fig:anatomy}
\end{figure}

\begin{figure}[t]
\centering
\includegraphics[width=0.62\linewidth]{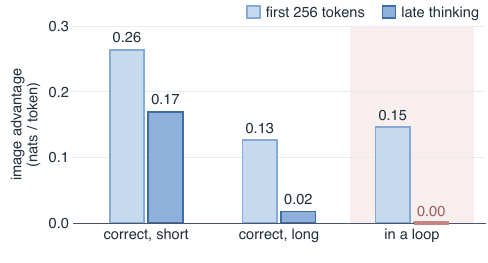}
\caption{\textbf{Inside a loop the image no longer matters.} Image advantage: per-token log-likelihood gain of the 0.8B
model's own reasoning trace with the real image instead of a gray one, from the editor's blank-image reads.}
\label{fig:image}
\end{figure}

\noindent\textbf{Independent editors return to the same layers.} The edits are legible, and independent editors place them in the same part
of the network (\cref{fig:anatomy}): most edits on both models change attention output projections in the middle
layers, and the three full-attention layers in the middle of the stack recur across editors and models. The 0.8B
\method{} edit removes, at half strength, a loop direction from the outputs of the attention and linear-attention blocks
of layers 8 to 22, one direction per layer. Its editor found the directions by contrasting the residual stream inside
repetitive thinking with token-matched stretches of normal thinking, and tested them causally before keeping them:
removing more reduced loops further, amplifying them made loops worse, and orthogonal directions of the same norm
did not reproduce the reduction (\cref{app:words08}). On the 2B model two editors reached the same part of the network by different routes: one
reflects a repetition direction in the outputs of four heads of two full-attention layers, the other strengthens a
single head of a third. Every edit \method{} selects from the original weights changes only attention output projections; none touches
the vision encoder or the MLPs.

\noindent\textbf{The repair adapts to the starting weights.} The prior edit already removes most loops but leaves
many answers cut at the token limit; started from it, the editors target these cut answers, and the selected edit
changes the MLP down projections of layers 15 and 17 to 20, including both layers the prior edit changed (\cref{app:blog}).

\noindent\textbf{The model drifts away from the image.} With blank-image reads, the 0.8B editor saw the image's
influence on the model's own tokens fade as thinking grows long and vanish inside a loop (\cref{fig:image}). The edit
makes this state rare.

\subsection{Ablation Study}
\label{sec:ablation}
\begin{figure}[t]
\centering
\includegraphics[width=0.62\linewidth]{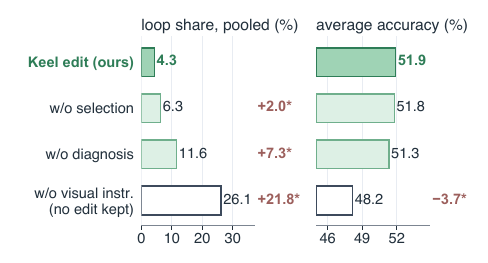}
\caption{\textbf{Every step of the edit is needed to remove the loops.} Each step of the edit (0.8B, original
weights) removed in turn; right of each panel: difference from the full edit ($^\ast$: 95\% interval excludes
zero). Lower loop share and higher accuracy are better.}
\label{fig:ablation}
\end{figure}

Every step of the \method{} edit is needed to remove the loops: removing any step from the edit of the original 0.8B
weights leaves significantly more answers ending in a loop (\cref{fig:ablation}; paired bootstrap, \cref{app:ablation}). Without the visual instruments, the text-only editor's edit fails the repair
test, so no edit is kept: the model keeps all its loops and is significantly less accurate. The prior edit
(\cref{tab:main}) also starts from the original weights, but its editor saw no images, worked on text-only
mathematics and had budgets of up to 50 passes~\cite{takehi2026weightspace}. Without the census and the repair
objective, editors judged by accuracy alone mostly commit no edit at all (\cref{app:editors}).
\section{Related Work}
\label{sec:related}

\noindent\textbf{LLMs as optimisers and weight editors.}
Large language models (LLMs) optimise prompts and programs~\cite{yang2023large,khattab2023dspy} and run machine-learning
experiments~\cite{lu2024the,chan2024mlebench}. LLM agents are also trained with reinforcement learning to call
tools~\cite{zheng2026procedureaware} and to work as voice agents~\cite{DBLP:journals/corr/abs-2608-26432}, and
multi-agent loops of vision-language models check and correct generated images over several
rounds~\cite{DBLP:journals/corr/abs-2602-06166}. A recent blog
post~\cite{takehi2026weightspace} has GPT-6 Astra read and write weights: it writes small
game-playing networks, fixes a control policy and removes a loop direction from the language model of
Qwen3.5-0.8B, an edit judged by accuracy and tested on text, mostly mathematics.
The editor never sees an image, and no image or GUI benchmark is evaluated. With the same editor,
\method{} is the first to carry such editing to VLMs on image tasks, from visual reasoning to GUI grounding: a census diagnoses
the failure, a repair objective judges each edit by the failure it removes, visual instruments show the editor the
images and where the VLM looks, and the harness selects among independent editors on questions none of them saw.
On 0.8B, \method{} starts from the blog's edit.

\noindent\textbf{Model editing and weight-space learning.}
Model editing changes weights to alter a specified behaviour: closed-form updates that rewrite stored
facts~\cite{meng2022locating,meng2022massediting}, arithmetic on task vectors~\cite{ilharco2022editing}, or
projecting a direction that mediates a behaviour out of the weights~\cite{arditi2024refusal}. Weights are also
combined without gradients, merging checkpoints of one reinforcement-learning
run~\cite{yang2026doesscalingreinforcementlearning} or specialist models~\cite{yang2026trainingfreeonpolicydistillation},
and pruning removes tokens or channels of vision transformers and vision-language-action models for faster
inference~\cite{DBLP:journals/pami/LiTMFCMWZ26,DBLP:conf/iclr/LiMSJTFMXWZ26}. Weight-space learning
trains networks that read or generate weights~\cite{han2026a,liang2025draganddrop}, and self-adapting models write
the data for their own fine-tuning~\cite{zweiger2025selfadapting}. Self-improving models learn from their own
outputs~\cite{fan2026audiorsirecursivelyselfimprovingaudio,DBLP:journals/tmlr/WangFGNJL26}. These methods must be told what to change, as a
fact, a task or a contrast set, or they need training; our editor is told only the symptom, finds where to act and
writes the fix without gradients.

\noindent\textbf{Automated interpretability.}
Mechanistic analyses localise behaviours to heads and directions~\cite{olsson2022incontext,gandelsman2023interpreting}
or read an audio LLM's answers out of its middle layers~\cite{DBLP:journals/corr/abs-2608-24958},
and agents now automate such analyses: MAIA~\cite{shaham2024a} finds spurious features in an image
classifier and fixes them by retraining a linear head on selected units. Our editor also experiments on a vision
model but writes its repair without training.

\noindent\textbf{Training-free fixes for VLMs.}
Hallucination and other visual errors are reduced by contrasting with distorted inputs~\cite{leng2023mitigating},
by penalising over-trust in summary tokens~\cite{huang2023opera}, by re-weighting attention towards the
image~\cite{liu2024paying,kang2025see,chen2025why,yin2025clearsight}, by steering hidden
states~\cite{liu2024reducing,chen2024ict}, or by projecting a hallucination subspace out of the
weights~\cite{yang2024nullu}. For reasoning models, steering vectors and component edits shorten, lengthen or
recalibrate thinking~\cite{chen2025seal,sun2025thinkedit,sun2026steeredit}, a lightweight scheduler decides when a
vision-language-action model thinks~\cite{DBLP:journals/corr/abs-2605-29438}, suppressing reflection words or forcing
a budget acts only on decoding~\cite{wang2025wait,muennighoff2025s}, what test-time computation gains is measured
against its full cost~\cite{yang2026languagemodelsgaintesttime}, and studies examine why reasoning
models loop~\cite{pipis2025wait,duan2026circular} and why long thinking weakens visual
grounding~\cite{liu2025more,tian2025more}. Each of these fixes applies a fixed recipe; our editor locates the cause
itself and may change almost any tensor. Training-based alternatives change the model with reinforcement
learning~\cite{DBLP:conf/icml/FanX22,DBLP:conf/iclr/FanZLH0ZWX23,DBLP:journals/corr/abs-2105-03923,DBLP:journals/corr/abs-2106-00707,DBLP:journals/corr/abs-2112-04145},
as in fine-tuning generative models~\cite{DBLP:conf/iclr/FanSCCLL25,DBLP:conf/nips/FanWCCL25,DBLP:journals/corr/abs-2510-18072},
rewarding the reasoning process~\cite{DBLP:conf/iclr/FanRLPSBGLG26} and training models to reason in fewer
tokens~\cite{DBLP:journals/corr/abs-2604-02322}; \method{} needs no training.

\noindent\textbf{Visual agents.} GUI grounding, the perception step of GUI agents, maps an instruction to a point
on a screenshot~\cite{cheng2024seeclick,wu2024osatlas}. We find that thinking loops cost both thinking models much
of their grounding accuracy, and that \method{}'s edits, written without any GUI data, significantly raise it.

\section{Conclusion}
\label{sec:conclusion}

Small reasoning VLMs such as Qwen3.5 lose many image answers to thinking loops, each spending the whole token budget, and the
remedies in use act on the decoding or must be told what to change. \method{} shows that a frontier model can find
the cause in the weights and remove it without training: editors trace the loops in the original
weights to the attention outputs of the middle layers, and selection keeps only edits that pass a repair test; inside
a loop, the smaller VLM stops consulting the image. The repaired models loop far less at the same size and per-token
cost and are significantly more accurate; on the best available weights, \method{} has the best average accuracy on
both models. A model's failure need not wait for retraining: agentic
weight editing joins decoding and training as a way to repair a model.

\bibliographystyle{unsrtcase}
\bibliography{main}

\clearpage
\appendix
\crefalias{section}{appendix}\crefalias{subsection}{subappendix}\crefalias{subsubsection}{subsubappendix}
\section{Discussion}
\label{app:discussion}

\subsection{What \method{} Solves}
The thinking mode of small vision-language models (VLMs) such as Qwen3.5 loses many answers to loops that run until the token limit. \method{} repairs that
failure by agentic weight editing: told only the symptom, the editor, a frontier multimodal LLM acting as an agent, locates the
cause, edits the weights and tests its edits within a small budget of forward passes, and the resulting edit removes
most loops, shortens the answers and makes the thinking model significantly more accurate on both models
(\cref{sec:main}). The edit is a change of a few tensors that a reader can inspect
(\cref{sec:analysis}), and the model it produces loads and runs like the original.

\clearpage
\subsection{Why Decoding-Time Remedies Are Not Enough}
The decoding-time remedies we compare leave the model unchanged. Budget forcing lets a loop run until the
budget and then forces an answer: on a large share of questions the answer is forced (\cref{app:loops}). The automated directional ablation needs the behaviour specified
in advance as a contrast between failing and correct answers; on the 2B model it finds no edit that passes its own
confirmation. The \method{} edit removes the loops in the model, so its answers are significantly shorter, and with budget
forcing significantly fewer answers have to be forced (\cref{app:loops}).

\clearpage
\subsection{Why a Repair Objective}
\label{app:objective}
\begin{figure}[!ht]
\centering
\includegraphics[width=\linewidth]{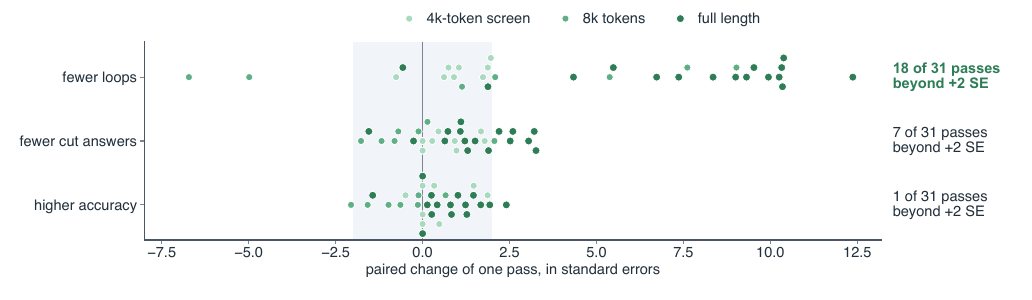}
\caption{\textbf{A repair is far easier to see than an accuracy change: loops fall by up to 12 standard errors.} Paired change of every paired generation pass of
the editor of the 0.8B \method{} edit (editor 7, \cref{tab:editor_passes}) in the loop share, the cut share and accuracy, in units of the
pass's own standard error and signed so that right is the repair direction; shaded: within two standard errors; marker colour: token limit of the pass (4{,}096, 8{,}192 or the full 16{,}384).}
\label{fig:snr}
\end{figure}

An accuracy change of a few points is hard to detect with the forward passes an editor can afford: on a batch of a
few hundred practice questions, its standard error is of the same order as the change itself. The failure the census
diagnoses is far easier to see: whether an answer is lost at the token limit, and whether it ends in a loop, is
observed on every question, and a working repair moves the share of looping answers far beyond its noise
(\cref{app:mech}) and with it the share of answers cut at the limit. The repair objective asks the editor to
reduce that cut share, which it can measure in every paired pass, and the selection step tests every final edit for it on
held-out questions. Accuracy enters as a constraint: the editor must not lower it significantly, and the selection step requires it to rise on the
held-out questions.
\Cref{fig:snr} shows the difference on the passes of the editor of the 0.8B \method{} edit (editor 7): 18 of its 31 paired generation passes
lower the loop share by more than two standard errors, by up to 12 standard errors, while accuracy rises by more than two standard
errors in 1 of them.

\clearpage
\subsection{Cost}
The only extra cost of \method{} is a pool of editing runs per model, three on the 0.8B model and four on the 2B model,
each several hours of an API model and a share of one GPU for its forward passes (\cref{app:compute}), plus one failure census
per model before editing (one generation pass on 512 practice questions, shared by all editors), one paired
held-out pass per edit for selection and one with budget forcing for the decoding choice; it is paid once. The edited model has the same size and the same per-token cost as the
original and needs no change to the inference code.

\FloatBarrier
\clearpage
\section{Algorithm Details}
\label{app:method}

\subsection{The Procedure and One Editing Session}
\label{app:procedure}
\Cref{alg:keel} gives the whole procedure of \cref{sec:method}, and \cref{alg:session} expands its inner loop,
the agent loop of one editing session: in each turn the editor reads the result of its last action and chooses the next. The harness, not the editor, runs every action.

\begin{algorithm}[h]
\caption{\method{} edit}
\label{alg:keel}
\footnotesize
\begin{algorithmic}[1]
\Require weights $\theta_0$, practice pool $\mathcal{D}_{\mathrm{prac}}$, its held-out part $H$, editors $1..K$, pass budget $B$
\State $C \gets$ census of $\theta_0$ on 512 practice questions \Comment{how the model fails}
\For{$k = 1, \dots, K$ \textbf{independently}}
  \State give editor $k$ the task, $C$, a kernel over $\theta_0$ and, optionally, the repair objective
  \State editor writes code for the kernel or requests forward passes until $B$ passes are used (or 260 turns or 14 hours)
  \State $\Delta_k \gets$ the submitted edit (possibly empty)
\EndFor
\State for each non-empty $\Delta_k$: $(a_k, c_k, s_k) \gets$ paired accuracy gain, change in cut share and its SE of $\theta_0{+}\Delta_k$ vs.\ $\theta_0$ on $H$
\State $S \gets \{k : c_k < -2 s_k,\ a_k > 0\}$; \textbf{if} $S = \emptyset$ \textbf{return} $\theta_0$
\State $k^\star \gets \arg\max_{k \in S} a_k$; score the union of the two best in $S$ if they touch disjoint tensors
\State \Return $\theta_0 + \Delta_{k^\star}$, or the union if it passes and is more accurate
\end{algorithmic}
\end{algorithm}

\begin{algorithm}[h]
\caption{One editing session of \method{}: the editor's agent loop}
\label{alg:session}
\footnotesize
\begin{algorithmic}[1]
\Require weights $\theta_0$, task statement $T$, census $C$, optional repair objective $O$, CPU kernel over $\theta_0$, pass budget $B$
\State $\Delta \gets \emptyset$; pass log $L \gets \emptyset$; $b \gets 0$
\While{$b < B$ \textbf{and} fewer than 260 turns and 14 hours have passed}
  \State the editor reads $T$, $C$, $O$ (if given), its notes, $L$ and the result of its last action (text and up to six images)
  \State the editor replies with an explanation and exactly one action
  \State code: run it in the kernel (inspect tensors of $\theta_0$ or $\theta_0{+}\Delta$, compute, save or drop $\Delta$)
  \State forward pass: run a generation pass on $\theta_0$, $\theta_0{+}\Delta$ or both, or a teacher-forced read on $\theta_0$ or $\theta_0{+}\Delta$; append a digest of its summary (its first lines) to $L$; $b \gets b+1$
  \State after 40 turns of a conversation, continue in a fresh one that receives $T$, $C$, $O$ (if given), the last 12{,}000 characters of the notes, $L$ and $\Delta$
\EndWhile
\State the editor makes its final choice; \Return the edit saved at that point (possibly empty)
\end{algorithmic}
\end{algorithm}

\begin{figure}[!ht]
\centering
\includegraphics[width=\linewidth]{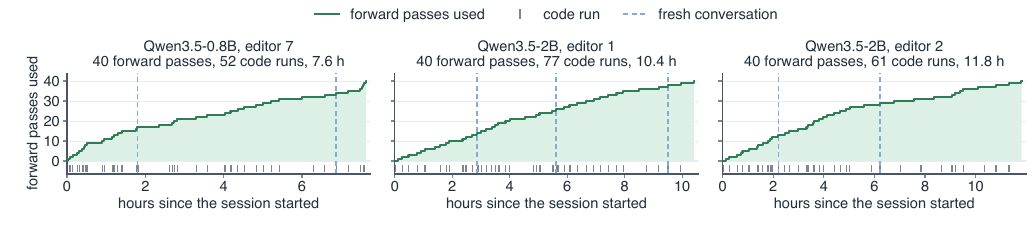}
\caption{\textbf{Each editing session is a long experiment that the editor directs itself.} The sessions that
produced the \method{} edits (numbered as in \cref{tab:editors}): forward passes used over time, the editor's code
runs (ticks) and the hand-offs to a fresh conversation (dashed).}
\label{fig:sessions}
\end{figure}

\Cref{fig:sessions} shows the three sessions that produced the \method{} edits.
Each alternates its own analysis code with forward passes until its 40 passes are spent, over 7.6 to 11.8 hours,
continuing in a fresh conversation two or three times.

\clearpage
\subsection{Editing Protocols}
\Cref{app:editors} lists every completed editing run that starts from the original weights. Protocol (a) gives the editor the
task and the inference protocol only, with the advice to judge candidates by accuracy. Protocol (b) adds the failure
census, states that repairing a diagnosed failure such as repetition is in scope, and reports the repetitive share of
cut answers in every generation summary. Protocol (c) adds the repair objective of \cref{sec:method} and the
paired difference in the share of cut answers in every paired generation summary. Visual editors have the instruments of
\cref{sec:method}; text-only editors run in the same harness without them.

\clearpage
\subsection{Harness Limits}
A session ends when its pass budget (40; 25 in protocol (a)) is used, after which the editor makes its final choice;
the editor cannot end it earlier. The harness also asks for the
final choice after 260 turns or 14 hours of wall time; two runs in \cref{app:editors} (with 38 passes) ended this
way. One code execution may take at most 900\,s and a session at most 300 executions; at most six images reach the editor
per turn and images older than two turns are dropped; full attention maps are returned for at most four texts per
read.

\clearpage
\subsection{Selection and Budget Forcing}
\label{app:select}
Every final edit is scored on the 1{,}000 held-out practice questions against the model the editors start from, with the same
per-question seeds, and selected by the repair test of \cref{sec:select} (\cref{alg:keel}). Budget forcing
is chosen on the same questions: with the kept edit, the model answers them in thinking mode with and without budget
forcing (same seeds), and \method{} uses budget forcing only if it is more accurate. It is more accurate on Qwen3.5-2B ($+3.5$
points, 95\% interval $[+1.0, +6.1]$) and not on Qwen3.5-0.8B ($-1.8$ points), so \method{} decodes with
budget forcing on the 2B model only.

\clearpage
\subsection{The 0.8B \method{} Edit in the Editor's Own Words}
\label{app:words08}
The box quotes two parts of the report that ends the editor's session, \emph{What changed} and \emph{Why this
edit}; [\dots] marks omitted text. The rest gives the editor's own validation on practice questions,
which selection re-checks on held-out questions.

\begin{tcolorbox}[colback=wsIce, colframe=wsNavy, boxrule=0.5pt, arc=2pt, left=4pt, right=4pt, top=3pt, bottom=3pt,
  fontupper=\footnotesize]
\textbf{What changed.} Only 15 attention output projections in zero-indexed language layers 8--22 were edited: 12
DeltaNet \texttt{out\_proj.weight} matrices and full-attention \texttt{o\_proj.weight} matrices in layers 11, 15, and
19. For each matrix, I applied $W' = W - 0.5\,u(u^\top W)$, where $u$ is a unit direction identified from
token-matched contrasts between repetitive thinking and nonrepetitive controls. This halves attention writes along
that direction while preserving orthogonal writes. MLPs, vision encoder, merger, normalization weights, embeddings,
and output head remain unchanged. No training was used.

\smallskip
\textbf{Why this edit.} [\dots] Broader causal tests localized the helpful intervention to attention outputs, particularly the
DeltaNet branch; suppressing the corresponding MLP writes instead worsened repetition. [\dots] Increasing suppression
progressively reduced repetitive failures; half-strength gave the best completion--accuracy balance. Amplifying the
same direction worsened repetition. A norm-matched orthogonal-direction control did not reproduce the strong
repetition reduction. [\dots]
\end{tcolorbox}

\clearpage
\subsection{The 2B \method{} Edit in the Editors' Own Words}
\label{app:2bwords}
The 2B \method{} edit is the union of the edits of editors 1 and 2 of the 2B model (\cref{tab:editors}). The boxes
quote what they changed and why; [\dots] marks omitted text. Both repair attention outputs, as the 0.8B
editor does, and report how the effect depends on the edit's strength and their own validation (\cref{tab:self_valid}).

\begin{tcolorbox}[colback=wsIce, colframe=wsNavy, boxrule=0.5pt, arc=2pt, left=4pt, right=4pt, top=3pt, bottom=3pt,
  fontupper=\footnotesize]
\textbf{Editor 1.} Retained the four-head edit [\dots] The main diagnosed failure was persistent repetition during
thinking. Activation analysis localized repetition-associated output directions in layer 11 heads 1--2 and layer 15
heads 6--7 (zero-indexed), using within-answer changes while controlling for changes seen in long, correct reasoning.
Each selected head received a rank-one reflection: $W_h' = W_h - 2(W_h d_h)d_h^\top$, where $d_h$ is its identified
unit direction. This reverses that coordinate while preserving orthogonal coordinates; it is not whole-tensor scaling.
[\dots] Directional intervention produced the predicted dose-response: repetitive cutoffs on the diagnostic set fell
from 25.0\% $\rightarrow$ 13.3\% $\rightarrow$ 7.1\% at strengths 0, 1, and 2. [\dots]
\end{tcolorbox}

\begin{tcolorbox}[colback=wsIce, colframe=wsNavy, boxrule=0.5pt, arc=2pt, left=4pt, right=4pt, top=3pt, bottom=3pt,
  fontupper=\footnotesize]
\textbf{Editor 2.} Retained one head-specific weight edit:\newline
\texttt{model.language\_model.layers.19.self\_attn.o\_proj.weight[:, 1024:1280] *= 1.5}\newline This strengthens
zero-indexed layer 19's attention head 4; all other columns and tensors remain unchanged [\dots] The dominant failure
was repetitive thinking that exhausted the token budget. I tested and rejected several correlated MLP neurons,
residual directions, and other attention heads before localizing a useful intervention: Head 4 at 0.5$\times$
worsened diagnostic accuracy by 7.81 points; 1.5$\times$ improved it by 7.81 points. [\dots] An equal-norm
amplification of neighboring head 5 worsened accuracy, supporting head specificity. [\dots] The best-supported
interpretation is preventing some trajectories from entering repetitive loops, not repairing visual perception inside
an established loop or forcing early thinking closure. [\dots]
\end{tcolorbox}

\begin{table}[!ht]
\centering
\small
\begin{tabular*}{\linewidth}{@{\extracolsep{\fill}}llrccc@{}}
\toprule
Model & Editor & Questions & Accuracy (\%)$\uparrow$ & Cut (\%)$\downarrow$ & Cut in a loop (\%)$\downarrow$ \\
\midrule
Qwen3.5-0.8B & 7 (\method{} edit) & 1{,}536 & 36.0 $\rightarrow$ \textbf{39.1} & 48.3 $\rightarrow$ \textbf{39.8} & 37.8 $\rightarrow$ \textbf{5.9} \\
Qwen3.5-2B & 1 (in the union) & 1{,}280 & 64.3 $\rightarrow$ \textbf{67.7} & 25.8 $\rightarrow$ \textbf{22.2} & 17.1 $\rightarrow$ \textbf{5.4} \\
Qwen3.5-2B & 2 (in the union) & 2{,}560 & 63.9 $\rightarrow$ \textbf{67.0} & 26.4 $\rightarrow$ \textbf{22.4} & 16.3 $\rightarrow$ \textbf{9.6} \\
\bottomrule
\end{tabular*}
\caption{\textbf{The editors of the \method{} edits validate their repairs themselves: higher accuracy, far fewer loops} (as reported: paired, full
token limit, practice questions not used to construct the edit; 0.8B: validation sets d--i of \cref{tab:editor_passes}).}
\label{tab:self_valid}
\end{table}

\FloatBarrier
\clearpage
\section{Experimental Details}
\label{app:setup}

\subsection{Models}
Qwen3.5-0.8B and Qwen3.5-2B~\cite{qwen3.5}, with the layout described in \cref{sec:setup}. Both are used as
released, in thinking mode, with the model card's chat template.

\clearpage
\subsection{Baselines}
\noindent\textbf{Budget forcing.} If the thinking has not closed after $16{,}384-1{,}024$ tokens, we append the
model's own thinking-budget sentence (\emph{``Considering the limited time by the user, I have to give the solution
based on the thinking directly now.''}) and the end-of-thinking token, and let the model answer in at most 1{,}024
tokens, so every method has the same total budget.

\noindent\textbf{Presence penalty.} The card's suggested remedy for endless repetition: presence penalty 1.5, other
sampling unchanged.

\noindent\textbf{Reflection suppression.} The token sequences of \emph{Wait}, \emph{wait}, \emph{Hmm}, \emph{hmm},
\emph{Alternatively}, \emph{alternatively}, \emph{However}, \emph{Maybe}, \emph{maybe}, \emph{Let me check},
\emph{double-check}, \emph{Double-check} and \emph{verify} are banned during decoding.

\noindent\textbf{Directional ablation.} The direction is the difference between the mean residual states (after a
chosen layer) of failing and of correct answers to 512 practice questions; it is projected out of every
attention-output, linear-attention-output and MLP-down matrix with strength $\alpha$. Layer and $\alpha$ are chosen
by paired accuracy on 256 practice questions per candidate, and the best candidate is kept only if its paired accuracy
gain on 256 fresh questions exceeds twice its standard error; \method{}'s repair test (\cref{app:select}) instead
asks that an edit lower the share of cut answers by more than two standard errors and raise accuracy on the 1{,}000
held-out questions. On the 0.8B model the baseline keeps layer 16 with $\alpha=0.5$; on the 2B model every layer it
tries lowers accuracy at full strength, and its best candidate (layer 10, $\alpha=0.5$) gains 2.7 points (SE 2.9) on
the fresh questions, below its own threshold, so the method returns the original weights.

\clearpage
\subsection{Benchmarks and Grading}
MathVista-mini (1{,}000 questions), MathVerse-mini vision-only (788), MMStar (1{,}500), HallusionBench (951), POPE
(1{,}500: a fixed random 500 per split), ScreenSpot-v2 (1{,}272). The answer is the last boxed expression after the
thinking; without one, the text after the last \emph{answer is} or \emph{answer:} (any case), and otherwise the last
line. Multiple-choice answers are matched by option letter (option text is mapped to its letter), integer answers
exactly, other numeric answers at the precision the benchmark states, free-form answers by mathematical equivalence,
yes/no answers by the first \emph{yes} or \emph{no} in the answer (else in the text after the thinking), and a ScreenSpot answer is correct if the predicted point lies
inside the target element's box.

\clearpage
\subsection{Inference}
vLLM with the model's chat template; thinking mode uses the model card's vision sampling (temperature 0.6, top-$p$
0.95, top-$k$ 20, presence penalty 0). Every question carries a fixed per-question seed, so paired comparisons sample
both models with the same seed. Images are resized to at most 1{,}310{,}720 pixels. The token limit is 16{,}384 for
every method.

\clearpage
\subsection{Practice Pool}
\label{app:pool}
ViRL39K~\cite{wang2025vlrethinker} without multi-image items, without items that repeat an evaluation question,
and without items that a 64-bit difference-hash screen (Hamming distance at most 4, compared only within buckets of equal
top 16 bits; ScreenSpot-v2 images not screened) flags as near-duplicates of an evaluation image. The census uses 512 practice questions; 1{,}000
questions are held out for selection and are never shown to an editor.

\clearpage
\subsection{Loop Detection}
An answer counts as cut if it reaches the token limit, and as ending in a verbatim loop if it is cut and its last
6{,}000 characters compress (zlib, level 6) to less than 15\% of their size.

\clearpage
\subsection{Statistical Tests}
All differences are paired over questions and tested with a bootstrap of 10{,}000 resamples. For the average over the
six benchmarks, each benchmark is resampled separately and the six means are averaged. Per-benchmark $p$-values are
Holm-adjusted over the six benchmarks. Shares and lengths pooled over the six benchmarks are resampled over all
questions.

\clearpage
\subsection{Hyperparameters}
\Cref{tab:hparams} collects every setting of the editing harness, the selection step and the evaluation.

\begin{table}[h]
\centering
\footnotesize
\setlength{\tabcolsep}{4pt}
\begin{tabular}{@{}p{0.43\columnwidth}p{0.53\columnwidth}@{}}
\toprule
Setting & Value \\
\midrule
\multicolumn{2}{@{}l}{\textit{Editor and harness}} \\
Editor & GPT-6 Astra, maximal reasoning effort \\
Forward-pass budget per session & 40 (protocol (a): 25) \\
Turns per conversation before a restart & 40 \\
Session end & budget used, or 260 turns, or 14\,h \\
Carried over at a restart & last 12{,}000 characters of notes \\
Generation pass & up to 512 questions (256 per model when paired); token limit up to 16{,}384, set by the editor \\
Teacher-forced read & up to 256 (question, answer) pairs \\
Code execution & at most 900\,s; at most 300 per session \\
Images per editor turn & at most 6; dropped after two turns \\
Attention maps per read & at most 4 texts \\
Out of scope & input embedding, output layer, final norm \\
\midrule
\multicolumn{2}{@{}l}{\textit{Census, objective and selection}} \\
Census & 512 practice questions \\
Repetitive ending & last 6{,}000 characters compress below 15\% \\
Held-out selection set & 1{,}000 practice questions \\
Repair test & paired change in cut share below $-2$ standard errors and in accuracy above 0 \\
Editors per pool & first round: 3 (0.8B), 4 (2B); second round: 6; on the prior edit: 3 \\
\midrule
\multicolumn{2}{@{}l}{\textit{Evaluation}} \\
Sampling (thinking mode) & $T{=}0.6$, top-$p$ 0.95, top-$k$ 20 \\
Token limit & 16{,}384 \\
Image size & at most 1{,}310{,}720 pixels \\
Samples per question & 1, fixed per-question seed \\
Budget forcing & thinking 15{,}360 + answer 1{,}024 tokens \\
Presence penalty baseline & 1.5 \\
Directional ablation & 512 questions (direction), 256 (search) \\
Bootstrap & 10{,}000 paired resamples; Holm over benchmarks \\
\bottomrule
\end{tabular}
\caption{\textbf{Hyperparameters} of the editing harness, the selection step and the evaluation.}
\label{tab:hparams}
\end{table}

\clearpage
\subsection{Computational Resources}
\label{app:compute}
All runs use NVIDIA RTX PRO 6000 Blackwell GPUs (96\,GB), shared between several editing runs through a forward-pass
server per GPU. \Cref{tab:cost} reports the cost of one editing run, averaged over the first-round runs of protocols (b) and (c).
Selection costs one paired pass over the held-out questions per edit. Runs that stop before submitting an edit are not
part of any pool (two 2B runs of protocol (c) and three second-round runs); one editor on the prior edit was resumed
from its saved state in a fresh conversation and finished its budget.
\Cref{tab:sessions} gives the same quantities for the three sessions that produced the \method{} edits (\cref{fig:sessions}).

\begin{table}[!htbp]
\centering
\small
\begin{tabular}{lr}
\toprule
Per editing run (mean over 12 runs) & \\
\midrule
Editor turns & 104 \\
Forward passes & 40 \\
Editor input tokens (mostly cached) & 3.9M \\
Editor output tokens & 168k \\
Selection passes per edit & 1 \\
\bottomrule
\end{tabular}

\caption{\textbf{Cost of one editing run} (first-round runs of protocols (b) and (c)).}
\label{tab:cost}
\end{table}

\begin{table}[!ht]
\centering
\footnotesize
\setlength{\tabcolsep}{3pt}
\begin{tabular*}{\linewidth}{@{\extracolsep{\fill}}lrrrrrrr@{}}
\toprule
Session & Turns & Code runs & Forward passes & Conversations & Hours & Editor input & Editor output \\
\midrule
Qwen3.5-0.8B, editor 7 & 93 & 52 & 40 & 3 & 7.6 & 3.45M & 155k \\
Qwen3.5-2B, editor 1 & 125 & 77 & 40 & 4 & 10.4 & 5.92M & 236k \\
Qwen3.5-2B, editor 2 & 105 & 61 & 40 & 3 & 11.8 & 3.73M & 168k \\
\bottomrule
\end{tabular*}

\caption{\textbf{Each session behind the \method{} edits finds its edit on its own in under 12 hours.} Turns: editor
replies (without the final report); code runs: code the harness executed; conversations: the session and its fresh
continuations; tokens: API counts.}
\label{tab:sessions}
\end{table}

\FloatBarrier
\clearpage
\section{Additional Experimental Results}
\label{app:results}

\subsection{The Failure Census}
\label{app:census}
\Cref{tab:census} gives the census that the editors of \method{}'s first-round pools receive at the start
(\cref{sec:method}), computed by
the harness on 512 practice questions under the inference protocol. On both models a large share of answers is cut at
the token limit and most of the cut answers end in repetitive text, while the answers that finish are mostly correct.
The census of the 0.8B model after the \method{} edit, which starts the second round (\cref{app:round2}), is the third
row of \cref{tab:census}: 12.8\% of its cut answers end in repetitive text.

\begin{table}[h]
\centering
\footnotesize
\setlength{\tabcolsep}{5pt}
\adjustbox{max width=\linewidth}{\begin{tabular}{lcccc}
\toprule
Model & Accuracy$\uparrow$ & Cut$\downarrow$ & Repetitive among cut$\downarrow$ & Accuracy when finished$\uparrow$ \\
\midrule
Qwen3.5-0.8B (thinking) & 34.2 & 46.9 & 84.6 & 64.3 \\
Qwen3.5-2B (thinking) & 62.1 & 29.3 & 70.7 & 87.8 \\
Qwen3.5-0.8B after the \method{} edit (start of the second round) & 40.8 & 38.3 & 12.8 & 66.1 \\
Qwen3.5-0.8B with the prior edit~\cite{takehi2026weightspace} & 40.4 & 34.6 & 9.0 & 61.8 \\
\bottomrule
\end{tabular}
}
\caption{\textbf{Failure census: after the \method{} edit, 12.8\% instead of 84.6\% of the cut 0.8B answers are repetitive.} On 512 practice questions (\%): accuracy, share of answers cut at the token limit,
share of the cut answers that end in repetitive text, and accuracy of the answers that finish.}
\label{tab:census}
\end{table}

\clearpage
\subsection{Significance Tests}
\label{app:tests}
\Cref{tab:sig_avg} compares \method{} (\cref{tab:main}) with every baseline evaluated on the same model, the prior edit
included, and the \method{} edit of the original weights with the thinking model it starts from.
\Cref{tab:sig_bench} gives the per-benchmark differences
between the \method{} edit and the thinking model. All twelve are positive: the edit raises accuracy on every benchmark of both models. On every benchmark of both models
it also leaves significantly fewer answers in loops and rescues significantly more answers from the token limit than it
newly cuts (\cref{tab:paired_len}).

\begin{table}[!htbp]
\centering
\small
\setlength{\tabcolsep}{6pt}
\begin{tabular}{llrr}
\toprule
Model & Comparison & Avg.\ diff.$\uparrow$ & 95\% interval \\
\midrule
Qwen3.5-0.8B & \method{} vs.\ Thinking & +7.92$^\ast$ & [+6.80, +9.05] \\
Qwen3.5-0.8B & \method{} vs.\ Presence penalty & +3.36$^\ast$ & [+2.25, +4.49] \\
Qwen3.5-0.8B & \method{} vs.\ Reflection suppression & +3.72$^\ast$ & [+2.60, +4.83] \\
Qwen3.5-0.8B & \method{} vs.\ Budget forcing & +2.56$^\ast$ & [+1.41, +3.70] \\
Qwen3.5-0.8B & \method{} vs.\ Directional ablation & +3.24$^\ast$ & [+2.12, +4.37] \\
Qwen3.5-0.8B & \method{} vs.\ Prior edit~\cite{takehi2026weightspace} & +1.07 & [$-$0.002, +2.13] \\
Qwen3.5-0.8B & \method{} vs.\ Prior edit + budget forcing & +0.56 & [$-$0.52, +1.62] \\
Qwen3.5-0.8B & \method{} edit vs.\ Thinking & +3.73$^\ast$ & [+2.68, +4.81] \\
\midrule
Qwen3.5-2B & \method{} vs.\ Thinking & +7.13$^\ast$ & [+6.12, +8.17] \\
Qwen3.5-2B & \method{} vs.\ Presence penalty & +2.03$^\ast$ & [+1.05, +3.05] \\
Qwen3.5-2B & \method{} vs.\ Reflection suppression & +3.64$^\ast$ & [+2.61, +4.69] \\
Qwen3.5-2B & \method{} vs.\ Budget forcing & +0.98$^\ast$ & [+0.04, +1.95] \\
Qwen3.5-2B & \method{} edit vs.\ Thinking & +5.22$^\ast$ & [+4.24, +6.25] \\
\bottomrule
\end{tabular}

\caption{\textbf{\method{}'s average accuracy is above every baseline's on both models} (0.8B: including the prior edit it starts from): paired difference in the average over the six benchmarks (points) and 95\% bootstrap interval; below, the
\method{} edit of the original weights against the thinking model it starts from. $^\ast$: interval
excludes zero.}
\label{tab:sig_avg}
\end{table}
\begin{table}[!htbp]
\centering
\small
\setlength{\tabcolsep}{4pt}
\begin{tabular}{lrrrrrr}
\toprule
Model & MathVista$\uparrow$ & MathVerse$\uparrow$ & MMStar$\uparrow$ & HallBench$\uparrow$ & POPE$\uparrow$ & ScreenSpot$\uparrow$ \\
\midrule
Qwen3.5-0.8B & +3.7$^\ast$ & +2.9 & +4.9$^\ast$ & +4.0$^\ast$ & +0.4 & +6.5$^\ast$ \\
Qwen3.5-2B & +2.0 & +2.5 & +3.2$^\ast$ & +4.7$^\ast$ & +1.8$^\ast$ & +17.1$^\ast$ \\
\bottomrule
\end{tabular}

\caption{\textbf{The \method{} edit raises accuracy on every benchmark of both models}: paired difference from the thinking model (points). $^\ast$: significant after Holm correction over the six
benchmarks ($p<0.05$).}
\label{tab:sig_bench}
\end{table}

\clearpage
\subsection{Every Editing Run}
\label{app:editors}
\begin{table}[!htbp]
\centering
\small
\begin{tabular}{lcccrrrrl}
\toprule
Model & Editor & Protocol & Visual & Passes & Weights & $\Delta$Acc$\uparrow$ (SE) & $\Delta$Cut$\downarrow$ (SE) & Outcome \\
\midrule
0.8B & 1 & (a) & \checkmark & 25 & 0.79M & +0.2 (1.5) & $-$6.2 (1.7) & passes \\
0.8B & -- & (a) & \checkmark & 25 & 0 & -- & -- & no edit \\
0.8B & -- & (a) & \checkmark & 25 & 0 & -- & -- & no edit \\
0.8B & -- & (a) & $\times$ & 25 & 0 & -- & -- & no edit \\
0.8B & -- & (a) & $\times$ & 25 & 0 & -- & -- & no edit \\
0.8B & 2 & (b) & \checkmark & 40 & 0.26M & +0.9 (1.6) & $-$7.2 (1.7) & passes \\
0.8B & 3 & (b) & \checkmark & 38 & 4.15M & +1.2 (1.5) & $-$8.0 (1.7) & passes \\
0.8B & -- & (b) & \checkmark & 40 & 0 & -- & -- & no edit \\
0.8B & 4 & (b) & $\times$ & 40 & 3,071 & +1.1 (1.4) & $-$4.6 (1.7) & passes \\
0.8B & 5 & (c) & \checkmark & 40 & 13.73M & +2.0 (1.6) & $-$6.7 (1.6) & passes \\
0.8B & 6 & (c) & \checkmark & 38 & 0.52M & +2.3 (1.6) & $-$6.2 (1.6) & passes \\
0.8B & 7 & (c) & \checkmark & 40 & 22.67M & +2.6 (1.7) & $-$8.7 (1.7) & selected \\
0.8B & 8 & (c) & $\times$ & 40 & 0.52M & $-$0.3 (1.6) & $-$7.1 (1.7) & fails \\
\midrule
2B & 1 & (b) & \checkmark & 40 & 1.94M & +1.5 (1.3) & $-$3.7 (1.4) & passes \\
2B & 2 & (b) & \checkmark & 40 & 0.52M & +2.9 (1.4) & $-$4.4 (1.3) & passes \\
2B & -- & (c) & \checkmark & 40 & 0 & -- & -- & no edit \\
2B & 3 & (c) & \checkmark & 40 & 0.26M & +1.2 (1.4) & $-$2.1 (1.4) & fails \\
2B & 1 + 2 & (b) & \checkmark & -- & 2.46M & +4.2 (1.4) & $-$4.0 (1.4) & selected \\
\bottomrule
\end{tabular}

\caption{\textbf{Every edit an editor commits lowers the share of cut answers on held-out questions.} Every completed editing run that starts from the original weights. Editor: number in
\cref{fig:anatomy} (runs that committed an edit); 1 + 2: the union of the two 2B edits, which selection keeps. Protocols: (a) task only, accuracy-judged; (b) with the failure census;
(c) with census and repair objective. Passes: forward passes used. Weights: number of changed weights.
$\Delta$Acc and $\Delta$Cut: paired change on the 1{,}000 held-out practice questions (points, standard error in
parentheses). Outcome of selection (\cref{sec:select}); the 0.8B pool of \method{} is protocol (c) with
visual instruments, the 2B pool all visual runs of protocols (b) and (c), because on 2B the (b) runs came first and the
completed (c) runs gave no passing edit. Applying the 2B pool rule to 0.8B would keep the same edit: editor 7 remains the most accurate
passing edit and shares the layer-15 full-attention output projection with editor 6, so no union is scored.}
\label{tab:editors}
\end{table}

\begin{figure}[!ht]
\centering
\includegraphics[width=\linewidth]{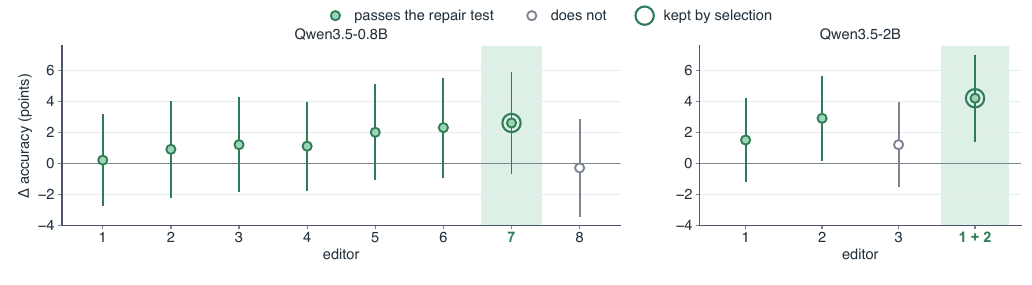}
\caption{\textbf{Selection keeps the most accurate edit on both models.} Held-out accuracy of every edit in
\cref{tab:editors}, numbered as there: paired change on the 1{,}000 held-out practice questions (points, 95\%
intervals). Filled: passes the repair test (cut share lowered by more than two standard errors and accuracy raised);
ringed and shaded: kept by selection. Higher is better.}
\label{fig:heldout}
\end{figure}

\Cref{tab:editors} lists the held-out score of every completed editing run that starts from the original weights,
numbered as in \cref{fig:anatomy}, and of the 2B union. With the
repair objective, every visual editor on the 0.8B model writes an edit that passes the repair test.
\Cref{fig:heldout} plots the held-out accuracy: on both models the edit that selection keeps is the most accurate
of all, and every edit that an editor commits also lowers the share of cut answers (\cref{tab:editors}).

\clearpage
\subsection{Paired Tests of the Ablation}
\label{app:ablation}
\begin{table}[!htbp]
\centering
\footnotesize
\setlength{\tabcolsep}{4.2pt}
\begin{tabular}{lccccc}
\toprule
& Diagnosis & Visual & Select. & Avg.$\uparrow$ & Loops$\downarrow$ \\
\midrule
Thinking &  &  &  & 48.2 & 26.1 \\
\midrule
w/o diagnosis & $\times$ & \checkmark & \checkmark & 51.3 & 11.6 \\
w/o visual instruments & \checkmark & $\times$ & \checkmark & 48.2 & 26.1 \\
w/o selection & \checkmark & \checkmark & $\times$ & 51.8 & 6.3 \\
\rowcolor{ours!12}\method{} edit & \checkmark & \checkmark & \checkmark & 51.9 & 4.3 \\
\bottomrule
\end{tabular}

\caption{\textbf{Every step of the edit is needed.} Ablation of the \method{} edit of the original 0.8B weights
(\checkmark\ kept, $\times$ removed), evaluated as in \cref{tab:main} without budget forcing. Diagnosis: census and
repair objective (without them, editors are judged by accuracy alone, budget of 25 forward passes); Visual: visual
instruments (the text-only edit fails the repair test, so no edit is kept); Select.: selection on held-out questions
(\emph{w/o selection}: mean of the three single runs of the pool). Avg.: average accuracy over the six benchmarks (\%);
Loops: answers ending in a verbatim loop, pooled over all questions (\%).}
\label{tab:ablation}
\end{table}

\Cref{tab:ablation_tests} gives, for every row of the ablation in \cref{tab:ablation} (\cref{fig:ablation}), the paired difference
from the \method{} edit in average accuracy and in the share of answers that end in a verbatim loop. Every ablated variant
leaves significantly more loops than the \method{} edit; without the visual instruments, where no edit passes the repair test,
the model is also significantly less accurate.

\begin{table}[h]
\centering
\footnotesize
\setlength{\tabcolsep}{3pt}
\begin{tabular}{lcc}
\toprule
Variant minus \method{} edit & Avg.\ accuracy$\uparrow$ & Loops$\downarrow$ \\
\midrule
w/o diagnosis & $-$0.58 [$-$1.70, +0.54] & +7.27$^\ast$ [+6.46, +8.10] \\
w/o visual instruments & $-$3.73$^\ast$ [$-$4.81, $-$2.68] & +21.84$^\ast$ [+20.80, +22.85] \\
w/o selection & $-$0.11 [$-$0.76, +0.53] & +2.02$^\ast$ [+1.63, +2.42] \\
\bottomrule
\end{tabular}

\caption{\textbf{Removing any step leaves significantly more loops}: paired differences from the \method{} edit on Qwen3.5-0.8B (points, 95\% bootstrap interval over
questions; average accuracy over the six benchmarks, loops pooled). $^\ast$: interval excludes zero.}
\label{tab:ablation_tests}
\end{table}

\clearpage
\subsection{Applying the Edit Step Again}
\label{app:round2}
The edit step can be applied to its own output. On the 0.8B model we ran a second round from the \method{} edit: a census
of the repaired model, six protocol-(c) visual editors that start from the first-round edit, and selection against
the repaired model on the same 1{,}000 held-out questions. \Cref{tab:round2_editors} lists their held-out scores; all
six pass, and the most accurate one is selected.

The selected second-round edit leaves every tensor of the first round unchanged and changes six numbers: it raises
the \texttt{A\_log} parameter of six linear-attention heads in layers 8, 9, 10 and 12 by $\ln 8$, which multiplies
their decay rate by eight, so these heads forget their running state faster. \Cref{tab:round2} shows the result. The
second round is significantly more accurate on average than the first and leaves significantly fewer answers in
loops (\cref{tab:round2_tests}).

\begin{table}[h]
\centering
\footnotesize
\setlength{\tabcolsep}{3pt}
\adjustbox{max width=\linewidth}{\begin{tabular}{lrrl}
\toprule
Second-round editor & $\Delta$Acc$\uparrow$ (SE) & $\Delta$Cut$\downarrow$ (SE) & Outcome \\
\midrule
a & $+$1.8 (1.5) & $-$7.4 (1.6) & passes \\
b & $+$3.2 (1.7) & $-$19.9 (1.6) & passes \\
c & $+$4.5 (1.6) & $-$5.8 (1.5) & selected \\
d & $+$4.4 (1.6) & $-$19.5 (1.6) & passes \\
e & $+$2.9 (1.6) & $-$32.8 (1.6) & passes \\
f & $+$2.7 (1.5) & $-$11.4 (1.5) & passes \\
union of c and d & $+$4.2 (1.6) & $-$24.6 (1.6) & passes, less accurate than c \\
\bottomrule
\end{tabular}
}
\caption{\textbf{All six second-round editors pass the repair test} on Qwen3.5-0.8B: paired change against the first-round \method{} edit on the
1{,}000 held-out practice questions (points, standard error in parentheses), and outcome of selection
(\cref{sec:select}).}
\label{tab:round2_editors}
\end{table}

\begin{table}[h]
\centering
\small
\setlength{\tabcolsep}{6pt}
\begin{tabular}{lc}
\toprule
Method & Avg.$\uparrow$ \\
\midrule
Thinking & 48.2 \\
\method{} edit (one round) & 51.9 \\
\rowcolor{ours!12}\method{} edit (two rounds) & 53.4 \\
\bottomrule
\end{tabular}

\caption{\textbf{A second round improves the edit further} on Qwen3.5-0.8B. Average accuracy over the six benchmarks (\%);
two rounds: the second-round edit selected on top of the first-round edit.}
\label{tab:round2}
\end{table}

\begin{table}[h]
\centering
\footnotesize
\setlength{\tabcolsep}{3pt}
\adjustbox{max width=\linewidth}{\begin{tabular}{lcc}
\toprule
Comparison & Avg.\ accuracy$\uparrow$ & Loops$\downarrow$ \\
\midrule
Two rounds vs.\ one round & +1.51$^\ast$ [+0.43, +2.60] & $-$1.30$^\ast$ [$-$1.87, $-$0.73] \\
\bottomrule
\end{tabular}
}
\caption{\textbf{The second round is significantly more accurate and leaves significantly fewer loops} on Qwen3.5-0.8B (points, 95\% bootstrap interval over questions;
average accuracy over the six benchmarks, loops pooled). $^\ast$: interval excludes zero.}
\label{tab:round2_tests}
\end{table}

\clearpage
\subsection{\method{} on Top of a Prior Edit}
\label{app:blog}
The blog post~\cite{takehi2026weightspace} released an edit of the
language model of Qwen3.5-0.8B, found on text-only mathematics; it changes the MLP down projections of layers 15 and
19. It is more accurate than the original weights on the 1{,}000 held-out questions (39.7\% vs.\ 36.2\%), so on the 0.8B model \method{}
starts from it: a census of that model
(\cref{tab:census}), three protocol-(c) visual editors that start from the prior edit, and selection against the
prior-edited model on the same 1{,}000 held-out questions (\cref{tab:blogstack_editors}). The census shows that the
prior edit already removes most repetitive endings, while a large share of answers is still cut at the token limit.
The selected edit changes the MLP down projections of layers 15 and 17 to 20.

The selected edit, applied together with the prior edit, is \method{} in \cref{tab:main}. It continues the repair
from the strongest available edit: the prior edit has removed most repetitive endings, but 34.6\% of the practice answers
are still cut at the token limit, only 9.0\% of them with a repetitive ending (\cref{tab:census}), and these cut answers are
what the repair objective asks the second edit to reduce; on the 1{,}000 held-out questions, the selected edit leaves 16.0
points fewer answers cut than the prior edit (SE 1.5, \cref{tab:blogstack_editors}). \method{} leaves significantly fewer answers in loops than the
prior edit ($-1.06$ points, 95\% interval $[-1.34, -0.78]$), writes significantly shorter answers than the prior edit with
budget forcing on every benchmark (\cref{tab:paired_len}), and, with budget forcing, reaches the budget on fewer answers on
every benchmark where the prior edit with budget forcing reaches it at all (\cref{tab:forced}). Its average accuracy is
higher than that of the prior edit, alone or with budget forcing (\cref{tab:sig_avg}). On Qwen3.5-2B, with no prior edit, \method{} starts from the original weights and is
significantly more accurate than every baseline (\cref{tab:sig_avg}).

\begin{table}[h]
\centering
\footnotesize
\setlength{\tabcolsep}{3pt}
\adjustbox{max width=\linewidth}{\begin{tabular}{lrrl}
\toprule
Editor on the prior edit & $\Delta$Acc$\uparrow$ (SE) & $\Delta$Cut$\downarrow$ (SE) & Outcome \\
\midrule
a & $+$1.4 (1.6) & $-$6.1 (1.5) & passes \\
b & $+$6.4 (1.5) & $-$12.1 (1.5) & passes \\
c & $+$7.6 (1.6) & $-$16.0 (1.5) & selected \\
\bottomrule
\end{tabular}
}
\caption{\textbf{Every editor on top of the prior edit passes the repair test} (0.8B): paired change against the prior-edited model on the
1{,}000 held-out practice questions (points, standard error in parentheses), and outcome of selection
(\cref{sec:select}).}
\label{tab:blogstack_editors}
\end{table}

\clearpage
\subsection{Loops and Budget Forcing per Benchmark}
\label{app:loops}
\begin{figure}[!htbp]
\centering
\begin{minipage}[t]{0.485\linewidth}
\centering
\includegraphics[width=\linewidth]{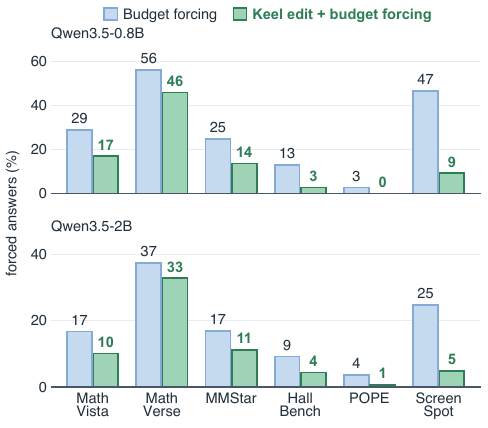}
\caption{\textbf{With the edit, budget forcing intervenes less often on every benchmark.} Share of answers whose
thinking reaches the budget and has to be forced to an answer, with budget forcing alone and with the \method{} edit
added. Lower is better.}
\label{fig:forced}
\end{minipage}\hfill
\begin{minipage}[t]{0.485\linewidth}
\centering
\includegraphics[width=\linewidth]{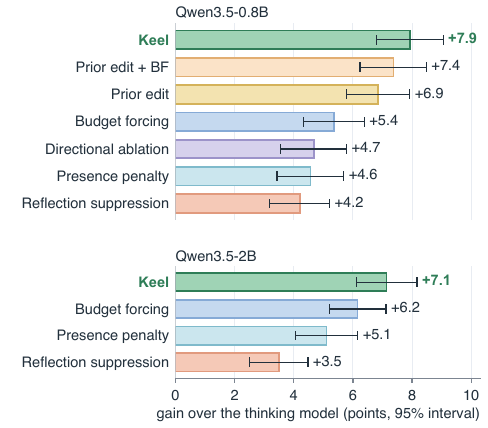}
\caption{\textbf{\method{} gains the most over the thinking model on both models.} Improvement of each method over
the released thinking model, from unrounded averages over the six benchmarks, with paired 95\% bootstrap intervals; BF: budget
forcing; prior edit: the edit released with~\cite{takehi2026weightspace} (0.8B only). On Qwen3.5-2B the directional
ablation kept no edit and is not shown. Higher is better.}
\label{fig:avg}
\end{minipage}
\end{figure}

\Cref{fig:loops} shows the loop shares per benchmark, and
\cref{tab:loops} reports for the thinking model, the \method{} edit and \method{} the share of cut answers, the share that end in a verbatim loop and the
mean number of generated tokens. With budget forcing, the edit reduces the share of answers that reach the thinking
budget and have to be forced to an answer on every benchmark and both models (\cref{fig:forced,tab:forced}).
\Cref{tab:paired_len} tests these differences. On every benchmark of both models, \method{}'s edit of the original
weights significantly reduces verbatim loops and, with budget forcing, the forced answers, and it rescues
significantly more answers from the token limit than it newly loses to it (\cref{fig:flow}); pooled over the
six benchmarks it significantly shortens the answers, alone and with budget forcing. Every per-benchmark difference in \cref{tab:paired_len} has the sign of the pooled one. Compared across benchmarks
(paired bootstrap of the difference between ScreenSpot-v2 and each other benchmark, Holm over the five), the edit
rescues significantly more answers on GUI grounding than on any other benchmark on both models, and on Qwen3.5-2B its
accuracy gain over the thinking model is significantly larger there than on any other benchmark.

\begin{table}[t]
\centering
\footnotesize
\setlength{\tabcolsep}{16.2pt}
\adjustbox{max width=\linewidth}{\begin{tabular}{lcccccc}
\toprule
Method & MathVista$\downarrow$ & MathVerse$\downarrow$ & MMStar$\downarrow$ & HallBench$\downarrow$ & POPE$\downarrow$ & ScreenSpot$\downarrow$ \\
\midrule
\multicolumn{7}{l}{\textbf{(a) Answers cut at the token limit, in parentheses those ending in a verbatim loop (\%)}} \\
\multicolumn{7}{l}{\textit{Qwen3.5-0.8B}} \\
Thinking & 28 (25) & 57 (52) & 27 (26) & 13 (13) & 2 (2) & 49 (49) \\
\method{} edit & 19 (6) & 47 (8) & 15 (4) & 3 (2) & 0 (0) & 9 (8) \\
\rowcolor{ours!12}\method{} & 7 (0) & 19 (1) & 3 (0) & 0 (0) & 0 (0) & 0 (0) \\
\multicolumn{7}{l}{\textit{Qwen3.5-2B}} \\
Thinking & 17 (13) & 36 (29) & 17 (14) & 9 (8) & 3 (3) & 28 (28) \\
\method{} edit & 12 (3) & 31 (3) & 10 (2) & 4 (2) & 1 (1) & 5 (5) \\
\rowcolor{ours!12}\method{} & 8 (0) & 25 (1) & 8 (0) & 2 (0) & 0 (0) & 0 (0) \\
\midrule
\multicolumn{7}{l}{\textbf{(b) Mean generated tokens per answer (thousands)}} \\
\multicolumn{7}{l}{\textit{Qwen3.5-0.8B}} \\
Thinking & 5.9 & 10.9 & 5.4 & 3.1 & 0.6 & 8.9 \\
\method{} edit & 5.5 & 10.5 & 4.5 & 2.2 & 0.3 & 4.8 \\
\rowcolor{ours!12}\method{} & 3.3 & 7.6 & 2.5 & 1.0 & 0.2 & 1.3 \\
\multicolumn{7}{l}{\textit{Qwen3.5-2B}} \\
Thinking & 4.5 & 8.7 & 4.2 & 2.8 & 0.8 & 5.9 \\
\method{} edit & 4.2 & 8.5 & 3.8 & 2.2 & 0.7 & 2.8 \\
\rowcolor{ours!12}\method{} & 3.9 & 8.4 & 3.8 & 2.3 & 0.6 & 2.8 \\
\bottomrule
\end{tabular}
}
\caption{\textbf{The edit leaves fewer answers cut and shorter answers on every benchmark.} Failure statistics in the
thinking mode, per benchmark: (a) share of answers cut at the token
limit and, in parentheses, of answers that are cut and end in a verbatim loop; (b) mean number of generated tokens.
\method{}: as in \cref{tab:main}.}
\label{tab:loops}
\end{table}

\begin{table}[t]
\centering
\footnotesize
\setlength{\tabcolsep}{2.5pt}
\begin{tabular*}{\linewidth}{@{\extracolsep{\fill}}llcccccc}
\toprule
Model & Method & MathVista$\downarrow$ & MathVerse$\downarrow$ & MMStar$\downarrow$ & HallBench$\downarrow$ & POPE$\downarrow$ & ScreenSpot$\downarrow$ \\
\midrule
Qwen3.5-0.8B & Budget forcing & 29 & 56 & 25 & 13 & 3 & 47 \\
Qwen3.5-0.8B & Prior edit + budget forcing & 11 & 31 & 6 & 1 & 0 & 0 \\
Qwen3.5-0.8B & \method{} + budget forcing & 6 & 17 & 3 & 0 & 0 & 0 \\
Qwen3.5-2B & Budget forcing & 17 & 37 & 17 & 9 & 4 & 25 \\
Qwen3.5-2B & \method{} & 10 & 33 & 11 & 4 & 1 & 5 \\
\bottomrule
\end{tabular*}

\caption{\textbf{With the edit, budget forcing has to force fewer answers on every benchmark.} Share of answers (\%)
whose thinking reaches the budget and is closed by the forcing sentence. \method{}: as in \cref{tab:main}; the \method{} edit with budget forcing: \cref{fig:forced}.}
\label{tab:forced}
\end{table}

\begin{table}[t]
\centering
\scriptsize
\setlength{\tabcolsep}{3.5pt}
\resizebox{\linewidth}{!}{\begin{tabular}{llrrrrrrc}
\toprule
Model & Comparison & MathVista & MathVerse & MMStar & HallBench & POPE & ScreenSpot & Pooled [95\% interval] \\
\midrule
\multicolumn{9}{l}{\textit{Mean generated tokens per answer (thousands)$\downarrow$}} \\
Qwen3.5-0.8B & \method{} edit vs.\ Thinking & $-$0.40 & $-$0.41 & $-$0.92$^\ast$ & $-$0.88$^\ast$ & $-$0.23$^\ast$ & $-$4.03$^\ast$ & $-$1.20$^\ast$ [$-$1.33, $-$1.06] \\
Qwen3.5-0.8B & \method{} edit + budget forcing vs.\ Budget forcing & $-$0.62$^\ast$ & $-$0.27 & $-$0.83$^\ast$ & $-$0.86$^\ast$ & $-$0.34$^\ast$ & $-$3.16$^\ast$ & $-$1.06$^\ast$ [$-$1.19, $-$0.92] \\
Qwen3.5-0.8B & \method{} vs.\ Prior edit + budget forcing & $-$1.42$^\ast$ & $-$1.99$^\ast$ & $-$1.23$^\ast$ & $-$0.72$^\ast$ & $-$0.04$^\ast$ & $-$1.06$^\ast$ & $-$0.99$^\ast$ [$-$1.06, $-$0.91] \\
Qwen3.5-2B & \method{} edit vs.\ Thinking & $-$0.31 & $-$0.22 & $-$0.45$^\ast$ & $-$0.61$^\ast$ & $-$0.19$^\ast$ & $-$3.01$^\ast$ & $-$0.84$^\ast$ [$-$0.95, $-$0.72] \\
Qwen3.5-2B & \method{} vs.\ Budget forcing & $-$0.53$^\ast$ & $-$0.50$^\ast$ & $-$0.47$^\ast$ & $-$0.52$^\ast$ & $-$0.41$^\ast$ & $-$2.35$^\ast$ & $-$0.82$^\ast$ [$-$0.93, $-$0.70] \\
\midrule
\multicolumn{9}{l}{\textit{Answers ending in a verbatim loop (percentage points)$\downarrow$}} \\
Qwen3.5-0.8B & \method{} edit vs.\ Thinking & $-$19.9$^\ast$ & $-$44.5$^\ast$ & $-$21.6$^\ast$ & $-$11.3$^\ast$ & $-$1.9$^\ast$ & $-$41.0$^\ast$ & $-$21.8$^\ast$ [$-$22.9, $-$20.8] \\
Qwen3.5-2B & \method{} edit vs.\ Thinking & $-$10.1$^\ast$ & $-$26.1$^\ast$ & $-$11.9$^\ast$ & $-$6.4$^\ast$ & $-$1.7$^\ast$ & $-$23.3$^\ast$ & $-$12.4$^\ast$ [$-$13.2, $-$11.5] \\
\midrule
\multicolumn{9}{l}{\textit{Answers forced by budget forcing (percentage points)$\downarrow$}} \\
Qwen3.5-0.8B & \method{} edit + budget forcing vs.\ Budget forcing & $-$11.9$^\ast$ & $-$10.4$^\ast$ & $-$11.1$^\ast$ & $-$10.3$^\ast$ & $-$2.7$^\ast$ & $-$37.3$^\ast$ & $-$14.0$^\ast$ [$-$15.0, $-$13.0] \\
Qwen3.5-2B & \method{} vs.\ Budget forcing & $-$6.6$^\ast$ & $-$4.6$^\ast$ & $-$5.7$^\ast$ & $-$4.8$^\ast$ & $-$3.0$^\ast$ & $-$19.8$^\ast$ & $-$7.6$^\ast$ [$-$8.5, $-$6.7] \\
\midrule
\multicolumn{9}{l}{\textit{Answers rescued from the token limit minus answers newly cut (percentage points)$\uparrow$}} \\
Qwen3.5-0.8B & \method{} edit vs.\ Thinking & +3.7$^\ast$ & +4.6$^\ast$ & +4.5$^\ast$ & +5.8$^\ast$ & +1.1$^\ast$ & +11.6$^\ast$ & +5.1$^\ast$ [+4.5, +5.8] \\
Qwen3.5-2B & \method{} edit vs.\ Thinking & +2.2$^\ast$ & +3.6$^\ast$ & +3.7$^\ast$ & +3.4$^\ast$ & +1.1$^\ast$ & +17.8$^\ast$ & +5.4$^\ast$ [+4.7, +6.1] \\
\bottomrule
\end{tabular}
}
\caption{\textbf{On every benchmark the edit shortens answers, leaves fewer loops, needs fewer forced answers and rescues more answers than it newly cuts}: edited minus reference, per benchmark
(tokens in thousands, shares in percentage points; $^\ast$: significant after Holm correction over the six
benchmarks) and pooled over all questions with a 95\% bootstrap interval ($^\ast$: interval excludes zero). Prior edit: the edit released
with~\cite{takehi2026weightspace}.}
\label{tab:paired_len}
\end{table}

\begin{figure}[!ht]
\centering
\includegraphics[width=\linewidth]{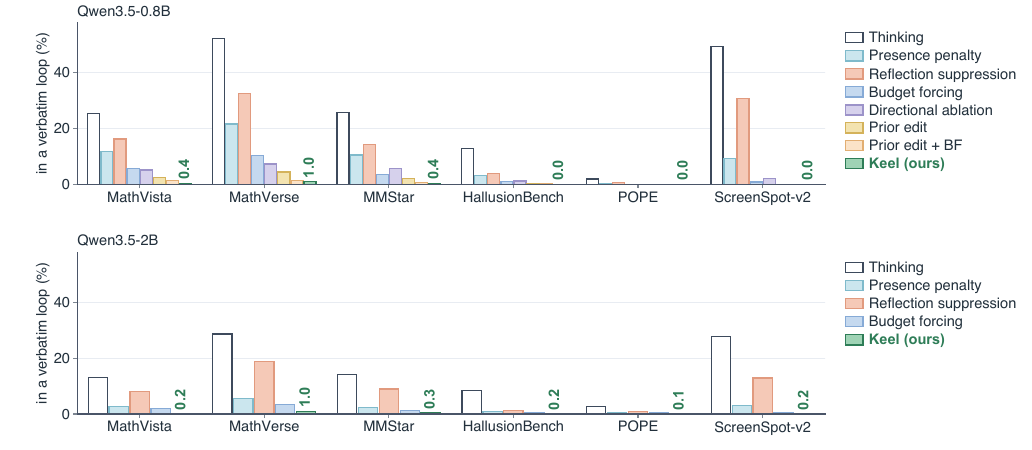}
\caption{\textbf{\method{} has the lowest or tied-lowest loop share on every benchmark.} Share of answers that
are cut at the token limit and end in a verbatim loop, for every method of \cref{tab:main}; the number above
the \method{} bar is its share. Prior edit: the edit released with~\cite{takehi2026weightspace}; BF: budget
forcing. Lower is better.}
\label{fig:loops_all}
\end{figure}

\begin{figure}[!ht]
\centering
\includegraphics[width=\linewidth]{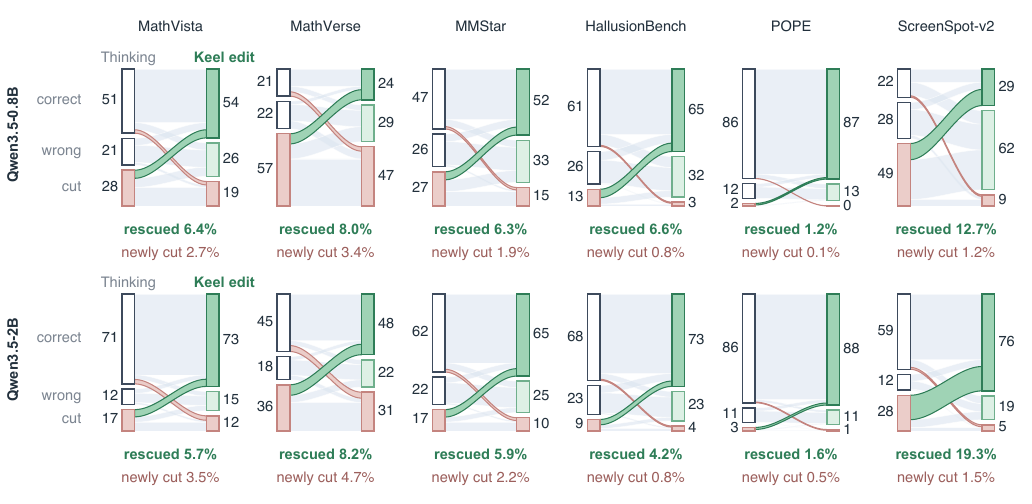}
\caption{\textbf{The edit rescues more answers from the token limit than it newly cuts on every benchmark.}
\Cref{fig:flow} per benchmark: outcome of the thinking model (left) and of the \method{} edit (right) on the
same question, as shares of the benchmark's questions (\%). Green: cut by the thinking model and answered correctly
after the edit (rescued); rose: answered correctly by the thinking model and cut after the edit (newly cut).}
\label{fig:flow_bench}
\end{figure}

\Cref{fig:loops_all} compares the loop share of every method of \cref{tab:main}: \method{} has the lowest
loop share, or ties for the lowest, on every benchmark of both models (\cref{tab:loops_all}). \Cref{fig:flow_bench} splits
\cref{fig:flow} by benchmark: on every benchmark of both models the edit rescues more answers from the token
limit than it newly cuts, most of all on GUI grounding (\cref{tab:transitions}).

\begin{table}[!ht]
\centering
\footnotesize
\setlength{\tabcolsep}{8.9pt}
\adjustbox{max width=\linewidth}{\begin{tabular}{lrrrrrrr}
\toprule
Method & MathVista$\downarrow$ & MathVerse$\downarrow$ & MMStar$\downarrow$ & HallBench$\downarrow$ & POPE$\downarrow$ & ScreenSpot$\downarrow$ & Avg.$\downarrow$ \\
\midrule
\multicolumn{8}{l}{\textit{Qwen3.5-0.8B}} \\
Thinking & 25.4 & 52.2 & 25.9 & 12.9 & 1.9 & 49.3 & 27.9 \\
Presence penalty & 11.7 & 21.6 & 10.5 & 3.3 & 0.1 & 9.3 & 9.4 \\
Reflection suppression & 16.2 & 32.5 & 14.3 & 4.0 & 0.5 & 30.7 & 16.4 \\
Budget forcing & 5.7 & 10.3 & 3.5 & 0.9 & \textbf{0.0} & 0.8 & 3.5 \\
Directional ablation & 5.1 & 7.2 & 5.5 & 1.2 & \textbf{0.0} & 2.1 & 3.5 \\
Prior edit~\cite{takehi2026weightspace} & 2.3 & 4.4 & 2.1 & 0.3 & \textbf{0.0} & \textbf{0.0} & 1.5 \\
Prior edit + budget forcing & 1.3 & 1.3 & 0.7 & 0.1 & \textbf{0.0} & \textbf{0.0} & 0.6 \\
\rowcolor{ours!12}\method{} (ours) & \textbf{0.4} & \textbf{1.0} & \textbf{0.4} & \textbf{0.0} & \textbf{0.0} & \textbf{0.0} & \textbf{0.3} \\
\midrule
\multicolumn{8}{l}{\textit{Qwen3.5-2B}} \\
Thinking & 13.0 & 28.7 & 14.1 & 8.4 & 2.7 & 27.9 & 15.8 \\
Presence penalty & 2.6 & 5.5 & 2.3 & 0.8 & 0.6 & 3.1 & 2.5 \\
Reflection suppression & 8.1 & 18.9 & 8.9 & 1.3 & 0.9 & 12.9 & 8.5 \\
Budget forcing & 2.0 & 3.4 & 1.3 & 0.4 & 0.5 & 0.6 & 1.4 \\
\rowcolor{ours!12}\method{} (ours) & \textbf{0.2} & \textbf{1.0} & \textbf{0.3} & \textbf{0.2} & \textbf{0.1} & \textbf{0.2} & \textbf{0.3} \\
\bottomrule
\end{tabular}
}
\caption{\textbf{\method{} has the lowest or tied-lowest loop share of every method on every benchmark} (\%, the numbers of \cref{fig:loops_all}): answers cut at
the token limit that end in a verbatim loop; lowest per column in bold. Avg.: mean over the six benchmarks.}
\label{tab:loops_all}
\end{table}

\begin{table}[!ht]
\centering
\footnotesize
\setlength{\tabcolsep}{4pt}
\begin{tabular*}{\linewidth}{@{\extracolsep{\fill}}llrrrrrrrrrr@{}}
\toprule
 & & & \multicolumn{3}{c}{Thinking: correct} & \multicolumn{3}{c}{Thinking: wrong} & \multicolumn{3}{c}{Thinking: cut} \\
\cmidrule(lr){4-6}\cmidrule(lr){7-9}\cmidrule(lr){10-12}
Model & Benchmark & $n$ & correct$\uparrow$ & wrong$\downarrow$ & cut$\downarrow$ & correct$\uparrow$ & wrong$\downarrow$ & cut$\downarrow$ & correct$\uparrow$ & wrong$\downarrow$ & cut$\downarrow$ \\
\midrule
Qwen3.5-0.8B & MathVista & 1{,}000 & 423 & 56 & 27 & 56 & 136 & 20 & \textbf{\color{ours}64} & 72 & 146 \\
 & MathVerse & 788 & 107 & 34 & 27 & 21 & 102 & 48 & \textbf{\color{ours}63} & 93 & 293 \\
 & MMStar & 1{,}500 & 597 & 84 & 28 & 90 & 276 & 21 & \textbf{\color{ours}95} & 140 & 169 \\
 & HallusionBench & 951 & 491 & 82 & 8 & 65 & 174 & 6 & \textbf{\color{ours}63} & 48 & 14 \\
 & POPE & 1{,}500 & 1{,}244 & 52 & 1 & 41 & 133 & 0 & \textbf{\color{ours}18} & 11 & 0 \\
 & ScreenSpot-v2 & 1{,}272 & 132 & 136 & 15 & 72 & 259 & 31 & \textbf{\color{ours}162} & 400 & 65 \\
\cmidrule(lr){2-12}
 & All six & 7{,}011 & 2{,}994 & 444 & 106 & 345 & 1{,}080 & 126 & \textbf{\color{ours}465} & 764 & 687 \\
\midrule
Qwen3.5-2B & MathVista & 1{,}000 & 638 & 36 & 35 & 34 & 74 & 15 & \textbf{\color{ours}57} & 42 & 69 \\
 & MathVerse & 788 & 290 & 30 & 37 & 22 & 95 & 27 & \textbf{\color{ours}65} & 45 & 177 \\
 & MMStar & 1{,}500 & 815 & 75 & 33 & 68 & 239 & 18 & \textbf{\color{ours}88} & 59 & 105 \\
 & HallusionBench & 951 & 597 & 43 & 8 & 56 & 151 & 10 & \textbf{\color{ours}40} & 27 & 19 \\
 & POPE & 1{,}500 & 1{,}251 & 28 & 8 & 39 & 131 & 2 & \textbf{\color{ours}24} & 11 & 6 \\
 & ScreenSpot-v2 & 1{,}272 & 659 & 77 & 19 & 68 & 81 & 10 & \textbf{\color{ours}245} & 81 & 32 \\
\cmidrule(lr){2-12}
 & All six & 7{,}011 & 4{,}250 & 289 & 140 & 287 & 771 & 82 & \textbf{\color{ours}519} & 265 & 408 \\
\bottomrule
\end{tabular*}

\caption{\textbf{On every benchmark the edit rescues more answers than it newly cuts}: outcome of every question before and after the edit (counts behind \cref{fig:flow} and
\cref{fig:flow_bench}). Column groups: the outcome of the thinking model; columns within a group: the outcome of the
\method{} edit on the same question; cut: reached the token limit without a correct answer. Green: rescued; newly cut:
correct $\rightarrow$ cut.}
\label{tab:transitions}
\end{table}

\clearpage
\subsection{Accuracy and Length}
\label{app:cost}
\Cref{fig:avg} gives the average gain of every method over the thinking model with paired intervals, and
\cref{fig:cost} plots average accuracy against the mean number of generated tokens. Pooled over the six benchmarks,
the \method{} edit generates significantly fewer tokens than the thinking model on both models, and \method{}
significantly fewer than the prior edit with budget forcing on the 0.8B model and than budget forcing on the 2B model
(\cref{tab:paired_len}).

\clearpage
\subsection{Inside the 0.8B \method{} Edit}
\label{app:mech}
\begin{figure}[!htbp]
\centering
\begin{minipage}[t]{0.485\linewidth}
\centering
\includegraphics[width=\linewidth]{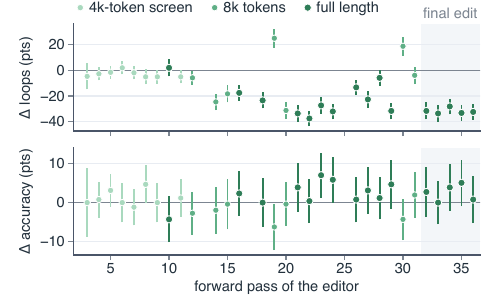}
\caption{\textbf{The editor's loop repairs are far beyond noise.} Each paired generation pass of the 0.8B \method{}
edit's editor, edited minus original (95\% interval). From pass 14 (a loop direction removed per middle layer), loops
move far beyond noise in almost every pass, a far clearer signal than single-pass accuracy (shaded: validation
passes of the final edit). Marker colour: token limit of the pass (4{,}096, 8{,}192 or the full 16{,}384). Fewer loops and higher accuracy are better.}
\label{fig:traj}
\end{minipage}\hfill
\begin{minipage}[t]{0.485\linewidth}
\centering
\includegraphics[width=\linewidth]{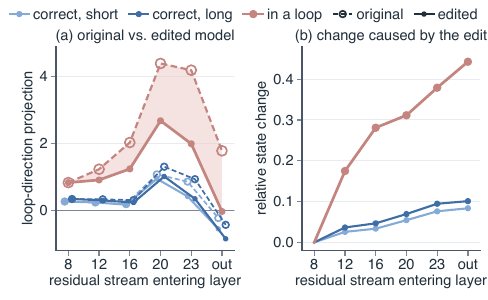}
\caption{\textbf{The edit removes the loop direction where loops form.} What the 0.8B \method{} edit does inside the network, read by its editor on 96 traces through
both models. (a) Projection of the residual stream entering layers 8, 12, 16, 20 and 23 (zero-indexed as in
\cref{sec:analysis}; out: after the last layer) on the loop direction measured there. (b) Relative change
of the residual state caused by the edit; the stream entering layer 8, the first edited layer, is unchanged.}
\label{fig:mech}
\end{minipage}
\end{figure}

\Cref{fig:traj} traces the paired generation passes of the editor of the 0.8B \method{} edit, from its first screens of single
neurons and one attention head to the final edit. This editor ended its session by reading the same traces through both models (\cref{fig:mech}). Inside
a loop the residual stream carries a large component along the loop directions, which the edit strongly reduces in
the upper layers, and the edit changes the hidden states inside loops several times more than during successful
reasoning. The editor's blank-image reads (\cref{fig:image}) show what the image contributes to the model's own
tokens: the real image makes them more likely at the start of the traces and during short successful reasoning, its
influence fades as thinking grows long, and inside a loop it vanishes. One 2B editor draws the same conclusion
explicitly: its edit keeps the model out of loops rather than repairing perception inside them
(\cref{app:2bwords}).

\begin{figure}[!ht]
\centering
\includegraphics[width=\linewidth]{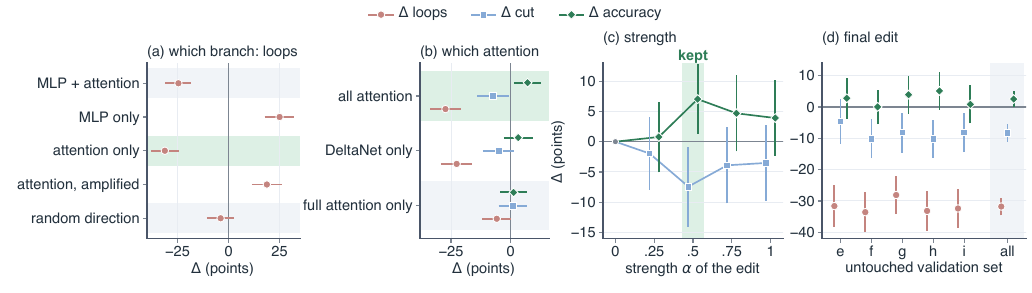}
\caption{\textbf{Each choice of the kept edit wins its test: only the attention writes carry the loops, both kinds of
attention together repair best, and half strength is the most accurate and cuts the fewest answers.} Passes of
editor 7 (0.8B \method{} edit) by what they tested; paired
change, edited minus original, 95\% intervals (lower is better except for accuracy); loops: cut in a verbatim loop; green band: the kept choice. (a)
Loop direction removed from the writes of layers 8--22 (8{,}192 tokens, strength 1; amplified: $-0.5$; random: same
norm, orthogonal). (b) Removed from all, from the linear-attention (DeltaNet) or from the full-attention writes (full
length, strength 0.5). (c) All attention writes, increasing strength. (d) The final edit on validation sets e--i
(256 unused questions each); all: pooled.}
\label{fig:editor_tests}
\end{figure}

\Cref{fig:editor_tests} groups the editor's generation passes by what they tested (all of them:
\cref{tab:editor_passes}). Removing the loop direction from the attention writes of layers 8--22 removes almost
every verbatim loop, while removing it from the MLP writes alone or amplifying the attention writes along it makes
loops more frequent, and a random direction of the same norm leaves them almost unchanged (a). The linear-attention
(DeltaNet) layers carry most of the effect, and editing the writes of both kinds of attention repairs best on all
three measures (b). Accuracy and the share of cut answers, by which the editor chose the strength, are both best at
half strength, the strength it kept (c). On the five validation sets that the editor ran on unused questions with
its final edit (sets e--i), it lowers the share of cut answers and removes most loops on every set; pooled over their 1{,}280
questions it cuts 8.3 points fewer answers, significantly (d). Both editors of the 2B edit report a dose response for their own heads
(\cref{app:2bwords}). \Cref{tab:editor_passes} also shows how editor 7 got there: it ruled out MLP
neurons, singly and as a set of twelve, loop directions in one and in four layers and one
linear-attention head
before removing the loop direction across layers 8--22, and from its first full-length validation of
that edit on, every validation on unused questions lowered the share of cut answers.

\begin{table}[!ht]
\centering
\scriptsize
\setlength{\tabcolsep}{5pt}
\adjustbox{max width=\linewidth}{\begin{tabular}{rlrrrrr}
\toprule
Pass & What the pass tested & Questions & Tokens & $\Delta$ accuracy$\uparrow$ & $\Delta$ cut$\downarrow$ & $\Delta$ loops$\downarrow$ \\
\midrule
1 & diagnosis: the original model on 32 practice questions & 32$\times$1 & 4{,}096 &  &  &  \\
3 & MLP neuron 11 of layer 19 kept at half strength & 32$\times$2 & 4{,}096 & 0.0 {\scriptsize (4.5)} & 0.0 {\scriptsize (5.5)} & $-$4.7 {\scriptsize (5.2)} \\
4 & the same neuron removed & 128$\times$2 & 4{,}096 & +0.8 {\scriptsize (2.4)} & $-$2.3 {\scriptsize (2.4)} & $-$2.7 {\scriptsize (2.6)} \\
5 & MLP neuron 16 of layer 13 removed & 128$\times$2 & 4{,}096 & +3.1 {\scriptsize (2.1)} & $-$3.9 {\scriptsize (2.3)} & $-$1.6 {\scriptsize (2.5)} \\
6 & 12 MLP neurons of layers 9--18 at half strength & 128$\times$2 & 4{,}096 & 0.0 {\scriptsize (2.6)} & $-$2.3 {\scriptsize (2.5)} & +2.0 {\scriptsize (2.6)} \\
7 & the same 12 neurons doubled & 128$\times$2 & 4{,}096 & $-$1.2 {\scriptsize (2.4)} & 0.0 {\scriptsize (2.7)} & $-$2.0 {\scriptsize (2.6)} \\
8 & loop direction removed from the MLP and linear-attention writes of layer 10 & 128$\times$2 & 4{,}096 & +4.7 {\scriptsize (2.5)} & $-$4.7 {\scriptsize (2.6)} & $-$5.1 {\scriptsize (2.9)} \\
9 & the same in layers 10, 12, 13 and 14 & 128$\times$2 & 4{,}096 & 0.0 {\scriptsize (2.6)} & $-$0.8 {\scriptsize (2.8)} & $-$5.1 {\scriptsize (2.7)} \\
10 & the layer-10 edit on untouched validation set a & 256$\times$1 & 16{,}384 & $-$4.3 {\scriptsize (3.0)} & +5.1 {\scriptsize (3.3)} & +2.0 {\scriptsize (3.4)} \\
11 & linear-attention head 7 of layer 1 at half strength & 128$\times$2 & 4{,}096 & +1.2 {\scriptsize (2.4)} & $-$1.2 {\scriptsize (2.6)} & $-$5.1 {\scriptsize (2.6)} \\
12 & the same head removed & 128$\times$2 & 8{,}192 & $-$2.7 {\scriptsize (2.8)} & +2.0 {\scriptsize (2.8)} & $-$5.9 {\scriptsize (2.8)} \\
14 & loop direction removed from MLP and attention writes, layers 8--22, $\alpha{=}1$ & 128$\times$2 & 8{,}192 & $-$2.0 {\scriptsize (3.1)} & +2.7 {\scriptsize (3.4)} & $-$24.6 {\scriptsize (3.2)} \\
15 & the same, $\alpha{=}0.5$ & 128$\times$2 & 8{,}192 & $-$0.4 {\scriptsize (3.3)} & +0.4 {\scriptsize (3.3)} & $-$18.4 {\scriptsize (3.4)} \\
16 & the $\alpha{=}0.5$ edit of layers 8--22 on untouched validation set b & 256$\times$1 & 16{,}384 & +2.3 {\scriptsize (2.9)} & $-$6.2 {\scriptsize (3.3)} & $-$17.6 {\scriptsize (3.2)} \\
18 & MLP and attention writes, $\alpha{=}1$ & 128$\times$2 & 16{,}384 & 0.0 {\scriptsize (3.2)} & $-$3.5 {\scriptsize (3.2)} & $-$23.4 {\scriptsize (3.2)} \\
19 & MLP writes only, $\alpha{=}1$ & 128$\times$2 & 8{,}192 & $-$6.2 {\scriptsize (3.0)} & +5.9 {\scriptsize (3.3)} & +25.0 {\scriptsize (3.7)} \\
20 & attention writes only, $\alpha{=}1$ & 128$\times$2 & 8{,}192 & $-$0.4 {\scriptsize (2.9)} & $-$0.4 {\scriptsize (2.8)} & $-$31.2 {\scriptsize (3.5)} \\
21 & attention writes only, $\alpha{=}1$ & 128$\times$2 & 16{,}384 & +3.9 {\scriptsize (3.2)} & $-$3.5 {\scriptsize (3.2)} & $-$33.6 {\scriptsize (3.4)} \\
22 & attention only, $\alpha{=}1$, on untouched validation set c & 256$\times$1 & 16{,}384 & +0.4 {\scriptsize (2.9)} & $-$2.3 {\scriptsize (3.2)} & $-$37.5 {\scriptsize (3.0)} \\
\rowcolor{ours!12}23 & attention writes only, $\alpha{=}0.5$ & 128$\times$2 & 16{,}384 & +7.0 {\scriptsize (2.9)} & $-$7.4 {\scriptsize (3.4)} & $-$27.3 {\scriptsize (3.3)} \\
\rowcolor{ours!12}24 & attention only, $\alpha{=}0.5$, on untouched validation set d (kept) & 256$\times$1 & 16{,}384 & +5.9 {\scriptsize (3.0)} & $-$9.4 {\scriptsize (3.1)} & $-$32.0 {\scriptsize (3.1)} \\
26 & attention writes only, $\alpha{=}0.25$ & 128$\times$2 & 16{,}384 & +0.8 {\scriptsize (3.0)} & $-$2.0 {\scriptsize (3.1)} & $-$13.3 {\scriptsize (3.1)} \\
27 & linear-attention (DeltaNet) writes only, $\alpha{=}0.5$ & 128$\times$2 & 16{,}384 & +3.1 {\scriptsize (3.1)} & $-$5.1 {\scriptsize (3.4)} & $-$22.7 {\scriptsize (3.4)} \\
28 & full-attention writes only (layers 11, 15, 19), $\alpha{=}0.5$ & 128$\times$2 & 16{,}384 & +1.2 {\scriptsize (2.8)} & +0.8 {\scriptsize (3.0)} & $-$5.9 {\scriptsize (3.1)} \\
29 & attention writes only, $\alpha{=}0.75$ & 128$\times$2 & 16{,}384 & +4.7 {\scriptsize (3.2)} & $-$3.9 {\scriptsize (3.2)} & $-$31.6 {\scriptsize (3.3)} \\
30 & attention loop writes amplified ($\alpha{=}{-}0.5$) & 128$\times$2 & 8{,}192 & $-$4.3 {\scriptsize (2.7)} & +3.9 {\scriptsize (3.3)} & +18.8 {\scriptsize (3.8)} \\
31 & random direction of the same norm, attention writes & 128$\times$2 & 8{,}192 & +2.0 {\scriptsize (3.0)} & $-$7.0 {\scriptsize (3.4)} & $-$3.9 {\scriptsize (3.4)} \\
\midrule
32 & final edit, untouched validation set e & 256$\times$1 & 16{,}384 & +2.7 {\scriptsize (3.3)} & $-$4.7 {\scriptsize (3.6)} & $-$31.6 {\scriptsize (3.4)} \\
33 & final edit, untouched validation set f & 256$\times$1 & 16{,}384 & 0.0 {\scriptsize (2.8)} & $-$10.2 {\scriptsize (3.2)} & $-$33.6 {\scriptsize (3.3)} \\
34 & final edit, untouched validation set g & 256$\times$1 & 16{,}384 & +3.9 {\scriptsize (3.1)} & $-$8.2 {\scriptsize (3.3)} & $-$28.1 {\scriptsize (3.1)} \\
35 & final edit, untouched validation set h & 256$\times$1 & 16{,}384 & +5.1 {\scriptsize (3.0)} & $-$10.2 {\scriptsize (3.1)} & $-$33.2 {\scriptsize (3.2)} \\
36 & final edit, untouched validation set i & 256$\times$1 & 16{,}384 & +0.8 {\scriptsize (3.1)} & $-$8.2 {\scriptsize (3.2)} & $-$32.4 {\scriptsize (3.1)} \\
\bottomrule
\end{tabular}
}
\caption{\textbf{The kept edit tests best on accuracy and cut answers together: before its final validation, its two passes
(attention writes only, {\boldmath$\alpha{=}0.5$}) raise accuracy more and cut fewer answers than any pass that tests another edit.}
Every generation pass of the editor of the 0.8B \method{} edit (editor 7) in order (the first is the diagnosis;
passes that only read states and likelihoods are not listed). Tests as in the editor's notes; loop direction: its rank-one
direction of the loop state, removed from the named writes with strength $\alpha$. Questions $\times$ samples (tuning:
128 tuning questions; validation: 256 unused practice questions). Paired change, edited minus original (points,
standard error in parentheses); loops: answers cut in a verbatim loop; shaded: the kept edit before its final validation.}
\label{tab:editor_passes}
\end{table}

\FloatBarrier
\clearpage
\section{Qualitative Examples}
\label{app:qual}
\begin{figure}[!ht]
\centering
\includegraphics[width=\linewidth]{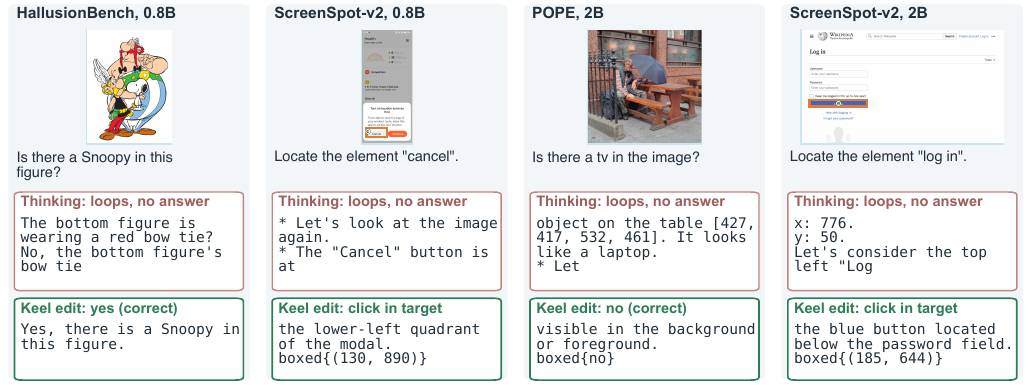}
\caption{\textbf{The edit stops the loops on the remaining benchmarks.} The remaining four picks of the rule of \cref{app:qual-rule}: the
benchmark--model pairs not shown in \cref{fig:qual,fig:qual2b}. ScreenSpot-v2: orange box, target element;
green ring, the click of the \method{} edit. Outputs as in \cref{app:qual-rule}.}
\label{fig:qual_extra}
\end{figure}

\subsection{Selection Rules}
\label{app:qual-rule}
\noindent\textbf{Examples in \cref{fig:qual,fig:qual2b,fig:qual_extra}.} For each benchmark and model, among the
questions with a single image on which the thinking model's evaluation answer was cut while ending in a verbatim loop
and the \method{} edit's evaluation answer was correct, we show the one with the shortest question text. We show the last
lines of the thinking model's answer and the end of the \method{} edit's answer; formulas are reduced to plain
text and long lines are wrapped.
\Cref{fig:gal_mathvista,fig:gal_mathverse_vo,fig:gal_mmstar,fig:gal_hallusion,fig:gal_pope,fig:gal_screenspot_v2}
continue the rule with the second and third shortest questions (ties broken by item identifier; the same ScreenSpot-v2
question is the pick for both models).

\noindent\textbf{Case study in \cref{fig:teaser}.} From the questions on which the thinking model's
evaluation answer was cut while repeating itself and the \method{} edit's answer was correct, we take the sixteen with the
shortest question text and let both models answer each of them again, three samples per question. Among the
questions on which the thinking model loops in all three samples and the \method{} edit is correct in all three, we show the
one with the shortest question. We applied the rule to MathVista and ScreenSpot-v2: on MathVista no candidate meets
the condition, on ScreenSpot-v2 exactly one does. We show the first of its three samples.

\noindent\textbf{Qwen3.5-2B and the remaining benchmarks.} On the larger model (\cref{fig:qual2b}) the thinking model checks the same point of a function graph, the
same pair of similar triangles, the same two leaves and the same circle construction until it is cut at the token
limit; after the edit, the model reaches an answer and closes its thinking.
\Cref{fig:qual_extra} completes the set: on HallusionBench, POPE and GUI grounding the thinking model circles over
the same figure, object or coordinates, and the edited model answers.

\clearpage
\subsection{More Examples on Every Benchmark}
\begin{figure}[!ht]
\centering
\includegraphics[width=\linewidth]{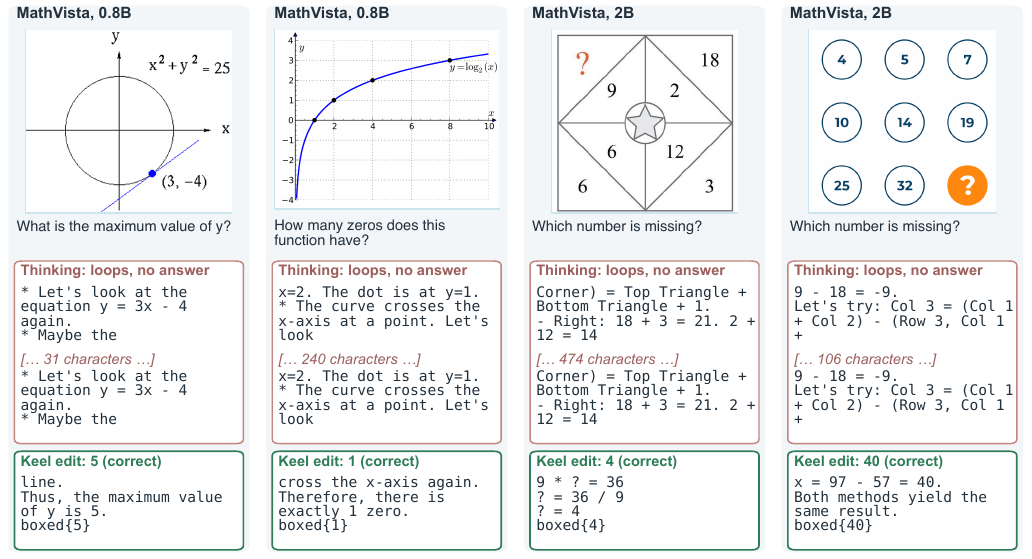}
\caption{\textbf{The edit ends the loop on more MathVista questions.} The second and third pick of the rule of
\cref{app:qual-rule} on each model. Rose box: the end of the thinking model's answer, cut at the token limit
(bottom), and the same lines where they last appeared (top), formulas as plain text; [\dots\ $n$ characters \dots]: the text
between them. Green box: the end of the \method{} edit's answer.}
\label{fig:gal_mathvista}
\end{figure}

\begin{figure}[!ht]
\centering
\includegraphics[width=\linewidth]{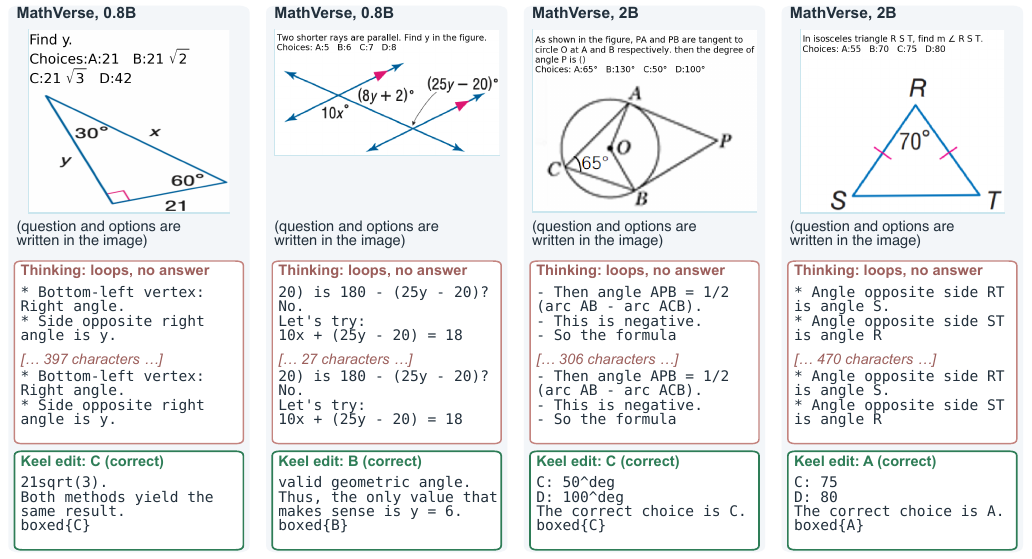}
\caption{\textbf{The edit ends the loop on more MathVerse questions.} The second and third pick of the rule of
\cref{app:qual-rule} on each model; boxes as in \cref{fig:gal_mathvista}.}
\label{fig:gal_mathverse_vo}
\end{figure}

\begin{figure}[!ht]
\centering
\includegraphics[width=\linewidth]{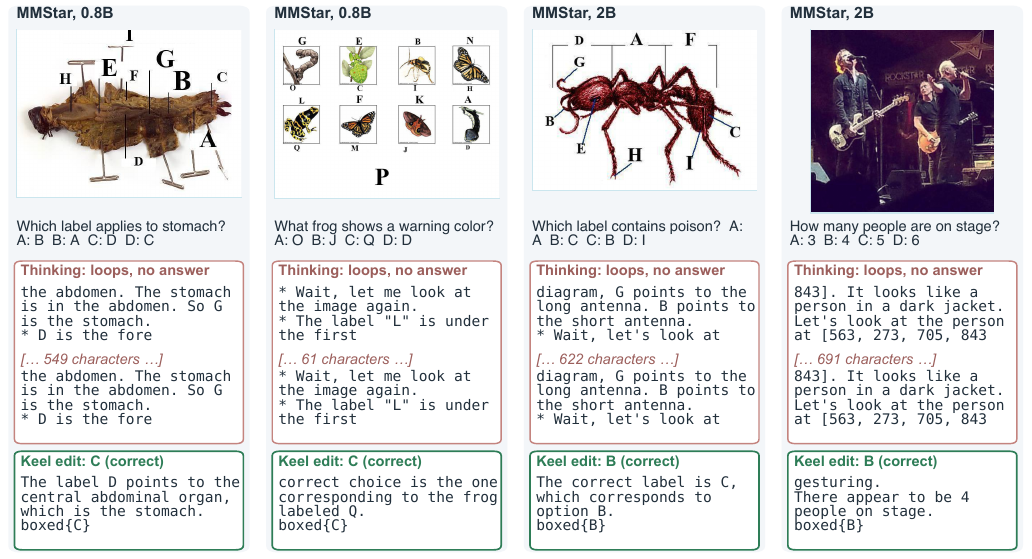}
\caption{\textbf{The edit ends the loop on more MMStar questions.} The second and third pick of the rule of
\cref{app:qual-rule} on each model; boxes as in \cref{fig:gal_mathvista}.}
\label{fig:gal_mmstar}
\end{figure}

\begin{figure}[!ht]
\centering
\includegraphics[width=\linewidth]{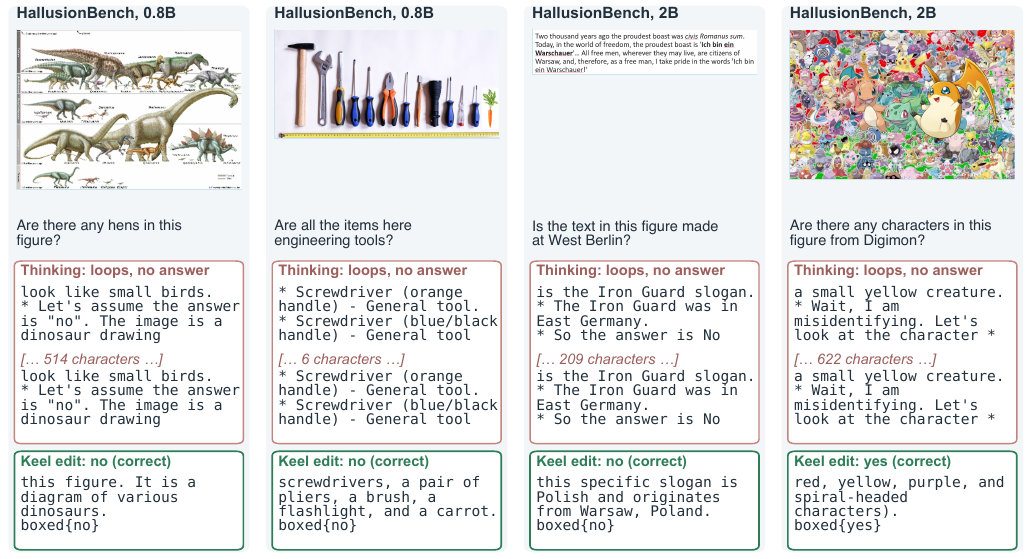}
\caption{\textbf{The edit ends the loop on more HallusionBench questions.} The second and third pick of the rule of
\cref{app:qual-rule} on each model; boxes as in \cref{fig:gal_mathvista}. In the first and third, the thinking
model names an answer, then keeps arguing until it is cut.}
\label{fig:gal_hallusion}
\end{figure}

\begin{figure}[!ht]
\centering
\includegraphics[width=\linewidth]{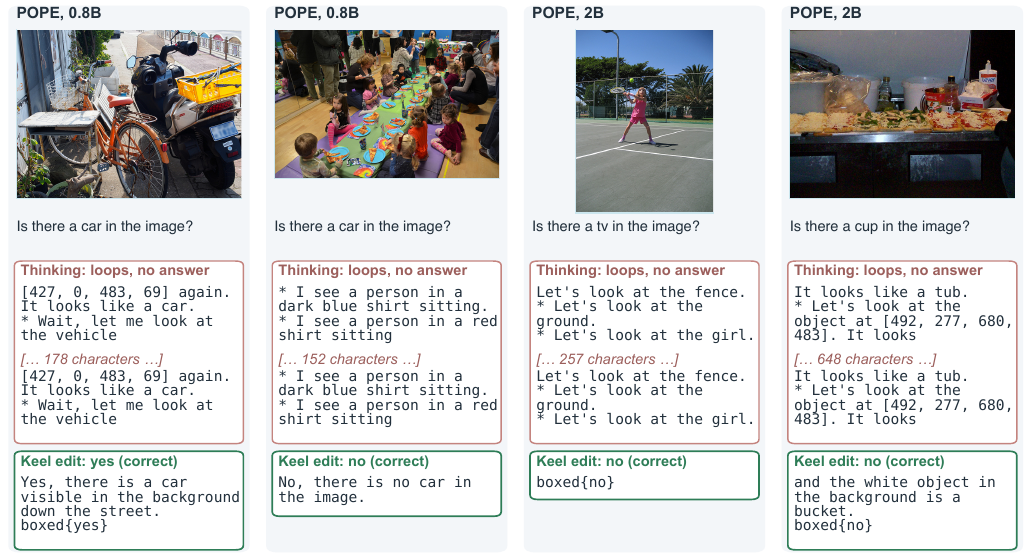}
\caption{\textbf{The edit ends the loop on more POPE questions.} The second and third pick of the rule of
\cref{app:qual-rule} on each model; boxes as in \cref{fig:gal_mathvista}.}
\label{fig:gal_pope}
\end{figure}

\begin{figure}[!ht]
\centering
\includegraphics[width=\linewidth]{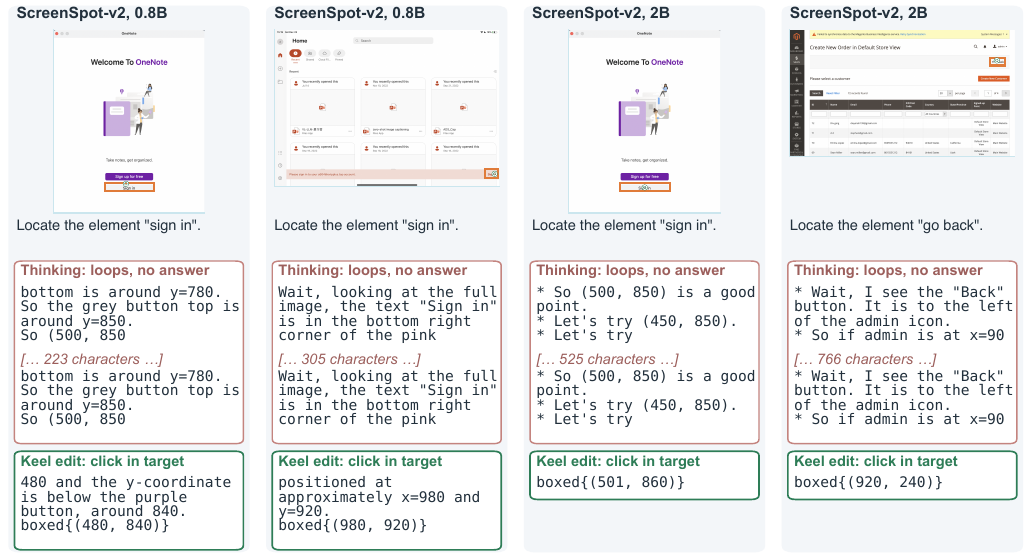}
\caption{\textbf{The edit ends the loop on more ScreenSpot-v2 questions.} The second and third pick of the rule of
\cref{app:qual-rule} on each model; boxes as in \cref{fig:gal_mathvista}. Orange box: target element; green ring: the click of the \method{} edit.}
\label{fig:gal_screenspot_v2}
\end{figure}

The repair is the same on every benchmark and both models (\crefrange{fig:gal_mathvista}{fig:gal_screenspot_v2}):
the thinking model repeats itself until it is cut after 16{,}384 tokens, and the edited model closes its thinking and
answers correctly after 146 to 12{,}909 tokens. The thinking models alternate between the same two remarks about a
line, return to the same dots of a logarithm curve or cycle through the same arithmetic rules of two number puzzles
(\textbf{MathVista}); re-derive the angles of the same right triangle, re-test the same angle equation, restate the
same tangent-angle formula or re-match the same sides of an isosceles triangle (\textbf{MathVerse}); re-read which
part a label points to, which label sits under which picture or which line leads to which antenna, or re-count the
same people on a stage (\textbf{MMStar}); re-inspect the small creatures of a dinosaur drawing, list the same two
screwdrivers, repeat the same historical argument or re-examine the same characters around a pair of wings
(\textbf{HallusionBench}); re-check the same vehicle, list the same seated people or re-scan the same objects of a
tennis court or a kitchen counter (\textbf{POPE}); and propose the same click again and again or re-locate the target
relative to the same neighbouring icons (\textbf{ScreenSpot-v2}).

\FloatBarrier

\end{document}